\documentclass[10pt,twocolumn,letterpaper]{article}

\usepackage[pagenumbers]{cvpr} % To force page numbers, \eg for an arXiv version

\usepackage{graphicx}
\usepackage{amsmath}
\usepackage{amssymb}
\usepackage{booktabs}
\usepackage{caption}
\usepackage{subcaption}
\usepackage{multirow}
\usepackage{tabularx}
\usepackage{soul}
\usepackage{color, xcolor}
\sethlcolor{pink}

\usepackage[accsupp]{axessibility}  % Improves PDF readability for those with disabilities.

\usepackage[pagebackref,breaklinks,colorlinks]{hyperref}

\usepackage[capitalize]{cleveref}
\crefname{section}{Sec.}{Secs.}
\Crefname{section}{Section}{Sections}
\Crefname{table}{Table}{Tables}
\crefname{table}{Tab.}{Tabs.}

\def\cvprPaperID{3030} % *** Enter the CVPR Paper ID here
\def\confName{CVPR}
\def\confYear{2022}

\begin{document}
	
	% \title{Learning a 4D Neural SMPL for Dynamic Human with Clothing and Hair}
	\title{H4D: Human 4D Modeling by Learning Neural Compositional Representation}
	
% 	\author{First Author\\
% 		Institution1\\
% 		Institution1 address\\
% 		{\tt\small firstauthor@i1.org}
% 		% For a paper whose authors are all at the same institution,
% 		% omit the following lines up until the closing ``}''.
% 		% Additional authors and addresses can be added with ``\and'',
% 		% just like the second author.
% 		% To save space, use either the email address or home page, not both
% 		\and
% 		Second Author\\
% 		Institution2\\
% 		First line of institution2 address\\
% 		{\tt\small secondauthor@i2.org}
% 	}
	
	\author{Boyan Jiang$^{1*}$ \quad Yinda Zhang$^{2*}$ \quad Xingkui Wei$^{1}$ \quad Xiangyang Xue$^{1}$ \quad Yanwei Fu$^{1}$ \\
$^{1}$Fudan University \quad $^{2}$Google
}

	\maketitle

{\let\thefootnote\relax\footnotetext{$^{*}$ indicates equal contributions.}}
{\let\thefootnote\relax\footnotetext{Boyan Jiang, Xingkui Wei and Xiangyang Xue are with the  School of Computer Science, Fudan University.}}
{\let\thefootnote\relax\footnotetext{Yanwei Fu is with the 
School of Data Science, Fudan University, and Fudan ISTBI—ZJNU Algorithm Centre for Brain-inspired Intelligence, Zhejiang Normal University, Jinhua, China.}}

	%%%%%%%%% ABSTRACT
	\begin{abstract}
		
		Despite the impressive results achieved by deep learning based 3D reconstruction, the techniques of directly learning to model 4D human captures with detailed geometry have been less studied. This work presents a novel framework that can effectively learn a compact and compositional representation for dynamic human by exploiting the human body prior from the widely used SMPL parametric model. Particularly, our representation, named H4D, represents a dynamic 3D human over a temporal span with the SMPL parameters of shape and initial pose, and latent codes encoding motion and auxiliary information. A simple yet effective linear motion model is proposed to provide a rough and regularized motion estimation, followed by per-frame compensation for pose and geometry details with the residual encoded in the auxiliary code. Technically, we introduce novel GRU-based architectures to facilitate learning and improve the representation capability. Extensive experiments demonstrate our method is not only efficacy in recovering dynamic human with accurate motion and detailed geometry, but also amenable to various 4D human related tasks, including motion retargeting, motion completion and future prediction. 
		Please check out the project page for video and code: \href{https://boyanjiang.github.io/H4D/}{https://boyanjiang.github.io/H4D/}.
		
	\end{abstract}
	
	%%%%%%%%% BODY TEXT
	\section{Introduction}
	\label{sec:intro}
	\vspace{-0.02in}
	
\begin{figure}[htb]
		\centering \includegraphics[width=0.95\linewidth]{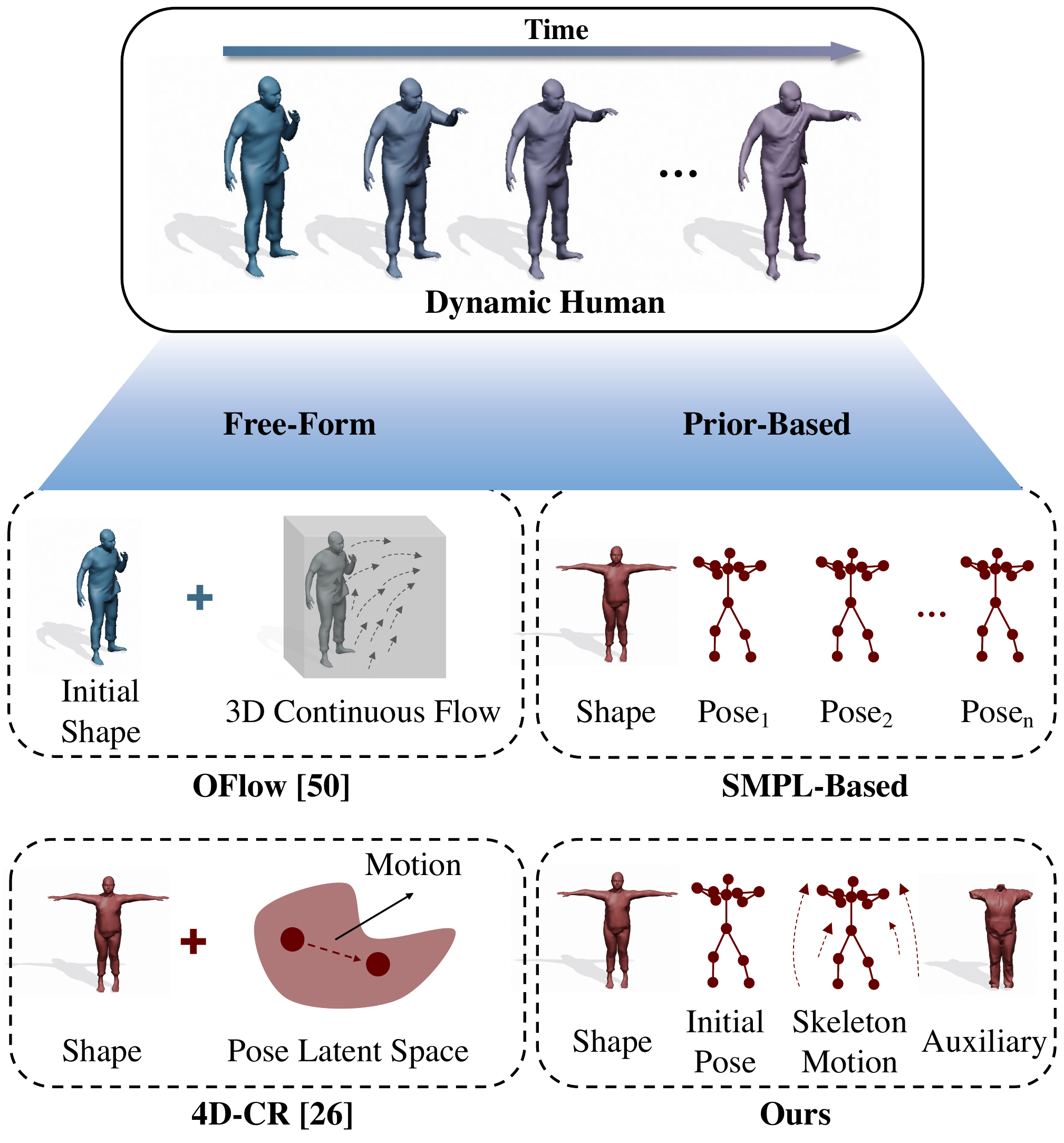}
		\vspace{-0.1in}
		\caption{\textbf{Comparison with existing 4D human representations.}
			Our representation supports faster inference and more complete reconstructions compared with free-form methods (Fig. \ref{fig:4d_recons}). And it provides long-range temporal context and additional fine-grained geometry controlled by low-dimensional SMPL parameters and latent codes, which is more compact compared with previous SMPL-based methods.
			\label{fig:teaser}}
		\vspace{-0.2in}
	\end{figure}

	The vanilla SMPL based parametric representations have been extensively studied and widely utilized for modeling 3D human shapes, and thus shown critical impacts to many human-centric tasks, such as pose estimation \cite{lassner2017unite,guler2019holopose,mehta2018single,kanazawa2019learning,kocabas2020vibe,choi2020beyond} and body shape fitting \cite{bogo2016keep,kolotouros2019learning,pavlakos2018learning,sengupta2021probabilistic,corona2021smplicit}.
	However, these representations are arguably insufficient for applications involving dynamic signals, \eg 3D moving humans (Fig.\ref{fig:teaser} top), since the temporal information is not captured.

	As solutions, 4D representations are proposed and can be in general categorized into free-form and prior-based methods depending on the 3D representation of the output shape (Fig. \ref{fig:teaser}).
 The free-form methods leveraging Neural ODE \cite{chen2018neural} 
 and deep implicit function \cite{niemeyer2019occupancy,jiang2021learning} often rely on computational expensive architectures to learn the compact latent spaces and reconstruct 4D sequences. Unfortunately, since the human body prior is not explicitly modeled, 
  the reconstruction results of these methods may contain obvious geometry artifacts such as missing hands, and their modeling errors accumulate rapidly over time. 
    On the other hand, prior-based methods \cite{kanazawa2019learning,zhang2019predicting,kocabas2020vibe} are mostly derived from the SMPL parametric model \cite{loper2015smpl}, which typically employs one shape parameter and a series of pose parameters to model dynamic sequences.
    Although they produce plausible results, their motion representations are not compact or only support a small time span, \eg $\pm$ 5 frames \cite{kanazawa2019learning}.

	In this paper, we propose \textbf{H4D}, which is a novel neural representation for \textbf{H}uman \textbf{4D} modeling that combines the merits of both the prior-based and free-form solutions.
	To reflect the compositional natures \cite{tokmakov2019learning}, we encode each dynamic human sequence with SMPL parameters representing shape and initial pose, and a compact latent code representing temporal motion, which can then be used to reconstruct the input sequence through a decoder.
	At the core of this decoder, a simple yet effective prior model extended from SMPL \cite{loper2015smpl} is designed to provide coarse but long-term estimation of the 3D human geometry and motion. This can ensure more complete and plausible outputs compared to the prior arts of free-form reconstructions \cite{niemeyer2019occupancy,jiang2021learning}, but potentially be inclined to suffer from the limited representation capability.
	To this end, we add an auxiliary latent code to compensate the inaccurate motion and enrich the geometry details.
	Such a representation takes full advantage of parametric models by exploiting strong prior based regularization for plausible initialization and complement them with powerful deep networks to facilitate the  human 4D modeling with impressive motion and geometry accuracy.

	Our representation is learned via an auto-encoding framework.
	The encoder predicts the SMPL parameters and latent codes for each aspect from densely sampled point clouds, which are fed into the decoder to reconstruct the identical input dynamic human sequence. 
	Once trained, the encoder and decoder are both fixed to support various applications, such as motion retargeting, completion, and prediction, through either forward propagation (feed-forward) or backward optimization (auto-decoding) depending on the inputs.
	We design novel Gated Recurrent Unit (GRU) \cite{cho2014learning} based architectures for both encoder and decoder to benefit the model performance while working in either mode.
	In feed-forward mode, we do not require the input point clouds to be temporarily tracked, \ie the point trajectories like in previous work \cite{niemeyer2019occupancy, jiang2021learning}. This simplifies the training requirements and enhances the applicability of high-level applications.
	In auto-decoding mode, our model leverages the temporal information for optimization, which is critical for robustness to recover detailed motion and geometry.

\noindent \textbf{Contributions}
	We propose H4D, a compact and compositional representation for 4D human captures, which combines a linear prior model with the residual encoded in a learned auxiliary code.
	The framework is learned via 4D reconstruction, and the latent representation can be extracted from either nonregistered point clouds in a feed-forward fashion or auto-decoding through optimization.
	Extensive experiments show that our representation and GRU-based architecture are effective in recovering accurate dynamic human sequences and providing robust performance for a variety of 4D human related applications, including motion retargeting/completion and future prediction.
	
	%-------------------------------------------------------------------------
	%------------------------------------------------------------------------
	\section{Related Work}
	\noindent \textbf{4D Representation} There has been a lot of work aiming to reconstruct 3D objects based on various representations, such as 3D voxels \cite{choy20163d,GirdharFRG16,wang2017cnn}, point clouds \cite{qi2016pointnet,fan2017point,QiLWSG18,achlioptas2017representation}, meshes \cite{groueix2018atlasnet,kanazawa2018learning,pixel2mesh,liao2018deep,pixel2mesh++} and implicit functions \cite{Occupancy_Networks, park2019deepsdf,jiang2020local,chibane2020implicit,erler2020points2surf,chabra2020deep}.
	However, the deep representation for 4D data, \textit{i.e.} a time-varying 3D object, has received less attention mostly due to the challenge of encoding the temporal dimension.
	Pioneer work mostly relies on Neural ODE \cite{chen2018neural} and combines with occupancy network \cite{Occupancy_Networks,niemeyer2019occupancy}, point clouds \cite{rempe2020caspr}, and compositional property \cite{jiang2021learning}.
	Despite state-of-the-art performance in various motion-related tasks, Neural ODE tends to accumulate errors over time that causes incomplete geometry, and slows down training convergence and inference run-time.
	In contrast, our model relies on the prior model for comprehensive geometry and motion and the recurrent network for efficient inference.
	
	\noindent \textbf{Human Body Estimation} 
	For human shape and pose estimation \cite{bogo2016keep,lassner2017unite,guler2019holopose,pavlakos2018learning,kolotouros2019learning,mehta2018single,kocabas2020vibe,choi2020beyond} or motion prediction \cite{martinez2017human,mao2019learning,barsoum2018hp,cai2020learning,aksan2019structured,aksan2020spatio}, most of works are based on SMPL  or its extension \cite{loper2015smpl,MANO:SIGGRAPHASIA:2017,SMPL-X:2019}.
	Specifically, HMMR \cite{kanazawa2019learning} learns to encode temporal information by reconstructing a small number of past and future frames. Zhang \etal \cite{zhang2019predicting} propose the first autoregressive model for predicting 3D human motion from image sequences. VIBE \cite{kocabas2020vibe} leverages GRU to regress SMPL parameters, and designs an adversarial learning framework to predict temporal transitions.
	Though produce plausible motion, the motion representations in these methods are either implicit \cite{kocabas2020vibe}, coupled with geometry \cite{kanazawa2019learning,zhang2019predicting,kocabas2020vibe}, or limited to a short temporal range \cite{kanazawa2019learning}.
	In contrast, we formulate the motion with a prior model based on PCA \cite{jolliffe2003principal,urtasun2006temporal} (its non-linear extension PGA \cite{fletcher2004principal} is also applicable) for long-range context followed by per-frame adjustments controlled by a learned latent code, which is compact, compositional, and tolerant to error accumulation.

	\noindent \textbf{Fine-grained Human Reconstruction} 
	Many human reconstruction methods \cite{bogo2016keep,kanazawa2018learning,kanazawa2019learning,kolotouros2019learning,mehta2018single,kocabas2020vibe} are limited to the unclothed body as SMPL-based models may suffer from  limited expressive power in the shape space.
	To capture fine-grained geometry, such as clothing or hair, the neural implicit function has been used to reconstruct a free-form surface \cite{saito2019pifu,saito2020pifuhd,chibane2020implicit,bhatnagar2020ipnet,zheng2021pamir,corona2021smplicit,saito2021scanimate,wang2021locally}, but it is still challenging to recover detailed structure like fingers, face, or wrinkles on clothes.
	Another family of approaches extends the parametric model by predicting per-vertex displacements upon the canonical body mesh \cite{alldieck2019learning,alldieck2018detailed,weng2019photo,alldieck2019tex2shape,lazova2019360,CAPE:CVPR:20}, which achieves a good balance between the expressiveness and prior regularization.
	Most related to us, CAPE \cite{CAPE:CVPR:20} trains a generator to synthesize fine-grained geometry from a latent space, and can run in the auto-decoding mode for fitting.
	However, it is empirically not robust to work with temporal frames and sensitive to the errors in imperfect poses, which is not ideal to plug and play in our scenario.
	
	\begin{figure*}[tb]
		\centering \includegraphics[width=0.92\linewidth]{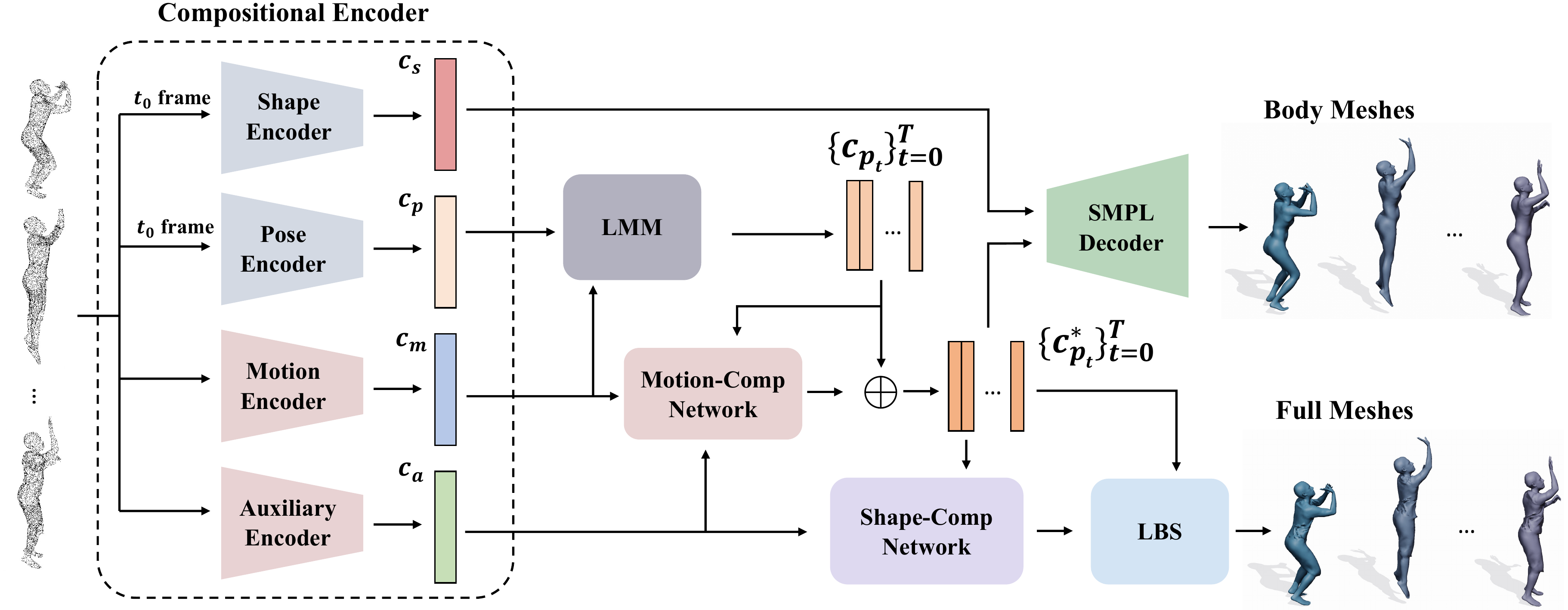}
		\vspace{-0.1in}
		\caption{\textbf{Overview of our framework.} We learn the compositional representation for dynamic humans through 4D reconstruction. Specifically, given an input point cloud sequence, we first extract SMPL parameters of shape and initial pose, and latent codes of motion and auxiliary information with the compositional encoder, and obtain a rough motion estimation with the Linear Motion Model (LMM).
		Then we predict the residuals of temporal motion and shape in canonical pose with the GRU-based Motion-Comp Network and Shape-Comp Network, respectively.
		Our method is able to output accurate body mesh sequences with the SMPL decoder \cite{loper2015smpl} as well as full mesh sequence with cloth and hair by using Linear Blend Skinning (LBS). Detailed architectures are in Supp. Mat.}
		
		\label{fig:pipeline}
		\vspace{-0.1in}
	\end{figure*}

	%-------------------------------------------------------------------------
	\section{Method}
	This section introduces our H4D representation, which is learned through reconstruction task (Fig.~\ref{fig:pipeline}).
	Given a 3D human model performing motion in a time span (mesh sequence of 30 frames), we sample a point cloud of 8192 points from each as the input sequence to the network. 
	Note that we do not assume the temporal correspondences among frames (\eg point trajectories) are available, which are critically required by previous 4D representations \cite{niemeyer2019occupancy,jiang2021learning}.
	
	The input sequence is fed into a compositional encoder to extract SMPL parameters representing shape and initial body pose, together with latent codes representing temporal motion and auxiliary of additional compensation on motion and geometry (Sec.~\ref{sec:encoder}).
	To reconstruct the input temporal sequence, the shape, initial pose, and motion codes are combined through a pre-learned Linear Motion Model (LMM) to generate a rough estimation of per-frame 3D shapes represented as SMPL~\cite{loper2015smpl} (Sec.~\ref{sec:linear_model}).
	Due to the limited capacity of LMM, the output, though plausible, demands additional refinements for accurate reconstruction.
	To this end, we feed the motion code, auxiliary code, and initial estimation to the GRU based Motion-Comp network (Sec.~\ref{sec:motion_comp}) and Shape-Comp network (Sec.~\ref{sec:shape_comp}) to predict the residual on temporal motion and shape in the canonical pose, respectively.
	The final sequence is obtained by deforming the refined canonical shape using the linear blending weight according to the refined per-frame poses.
	
	\subsection{Compositional Encoder} \label{sec:encoder}
	To keep the representation compositional, we train four separate encoders to extract SMPL parameters representing the shape $c_{s}$ and initial pose $c_{p}$, and latent codes representing motion $c_{m}$ and auxiliary information $c_{a}$.
	The shape and pose encoders are implemented as PointNet-based \cite{qi2016pointnet} network with ResNet blocks, which take only the starting frame as input since it is sufficient to tell the canonical body shape and initial pose.
	On the other hand, the motion and auxiliary encoders take all frames as input since temporal information is needed.
	To achieve that, we firstly encode the point cloud of each frame into a feature vector using a shallow PointNet, and then further aggregate per-frame feature with a GRU layer.
	The feature extractor is shared between motion and auxiliary encoders, and only GRUs are trained respectively.
	Note that our temporal encoders can process sequences of unordered point clouds without temporal correspondences.
% 	do not require the temporal correspondences of input, \ie point trajectories, and thus can process sequences with unordered point clouds. 

	\subsection{Linear Motion Model} \label{sec:linear_model}
	We take the predicted $c_{p}$ and $c_{m}$ to reconstruct a coarse estimation of motion.
	Inspired by Urtasun \etal \cite{urtasun2006temporal}, we employ the parameter space of SMPL model and pre-learn a linear model for the motion to ensure robustness.
	Each input temporal sequence can be represented as $\Phi=[ \theta_{1,}\ldots\theta_{L}], L=30$, where $\theta_i\in\mathbb{R}^{72}$ is the SMPL pose parameter for frame $i$.
	We then represent motion as the per-frame difference of the pose parameter from the first frame, i.e. $\Psi=[\theta_{2}-\theta_{1},\ldots,\theta_{L}-\theta_{1}]\in\mathbb{R}^{72(L-1)}$, and run a Principal Component Analysis (PCA) \cite{jolliffe2003principal} to reduce the dimension. 
% 	Note that the Principal Geodesic Analysis (PGA) \cite{fletcher2004principal} is also applicable to manifold-valued data in this case, but we found PCA is efficient enough to offer an initial estimation.
	The input motion now can be reconstructed through the linear model: $\hat{\Phi}=[\theta_{1},\alpha^{T}\cdot\mathcal{M}+\mu_{\Psi}+\theta_{1}]$,
% 	\begin{equation}\label{eq:delta}
% 	\hat{\Phi}=[\theta_{1},\alpha^{T}\cdot\mathcal{M}+\mu_{\Psi}+\theta_{1}]
% 	\end{equation}
	where $\alpha\in \mathbb{R}^{K}$ is the coefficient of principal components, $\mathcal{M}=\left[\mathbf{M}_{1},\ldots,\mathbf{M}_{K}\right]\in\mathbb{R}^{72(L-1)\times K}$ and $\mu_{\Psi}$ are the top $K$ principal components and mean of $\Psi$.
	
	In practice, we found it more robust to run PCA separately for the global orientation (\ie pelvis) and body joint rotation.
	We pick 4 bases for global orientation and 86 basis for body joint rotation, which explains $90\%$ of the variance\footnote{Please refer to Supp. Mat. for visualization of principal components}.
	Finally, we plug the linear motion model into our pipeline, amenable to the output of the compositional encoder: $\left\{{c_{p_t}}\right\}_{t=0}^{L-1}=[c_p,c_m^{T}\cdot\mathcal{M}+\mu_{\Psi}+c_p]$,
% 	\begin{equation}\label{eq:LMM}
% 	\left\{{c_{p_t}}\right\}_{t=0}^{L-1}=[c_p,c_m^{T}\cdot\mathcal{M}+\mu_{\Psi}+c_p],
% 	\end{equation}
	where $c_{p_t}$ is the pose parameter for frame $t$.

	\subsection{Motion Compensation Network} \label{sec:motion_comp}
	The LMM is effective in representing motion with a relatively large number of temporal frames; unfortunately, it lacks the capacity to represent motion details.
	As a result, the predicted pose sequences are not accurate enough. 
	
	To improve the motion accuracy, we build a motion compensation network (Motion-Comp) to adjust the pose parameter of each frame.
	Specifically, we adopt a GRU-based network \cite{cho2014learning} as it was demonstrated to be effective for processing temporal information.
	We concatenate the motion code $c_m$ and the auxiliary code $c_a$ to the pose parameters of each frame from LMM prediction $\left\{{c_{p_t}}\right\}_{t=0}^{L-1}$, and then feed them sequentially into  GRU to produce the residual of each frame.
	Once the per-frame pose parameters are updated with the output of Motion-Comp network, we combine them with the shape parameter $c_s$ from the encoder into the standard SMPL decoder to reconstruct the per-frame mesh. 
	Overall, our motion model benefits from both the strong prior in the linear motion model and the impressive capacity of the motion compensation network.
	
	\subsection{Shape Compensation Network \label{sec:shape_comp}}
% 	\vspace{-0.1in}
	So far, we are able to reconstruct the correct motion sequences, which can be further converted to body mesh sequences via SMPL decoder.
	However, the predicted shapes are still inferior, as many details such as hairs or clothes are missing. This is mostly due to the constrained capacity of SMPL shape space.
	To enhance the geometry, the shape representation presented in  CAPE~\cite{CAPE:CVPR:20} is introduced:  a per-vertex offset is estimated for the body mesh in the canonical space via a graph-based neural network conditioned on target pose.
	The added details would be then transferred to the target body pose via the pre-defined linear blending weight in SMPL.
	When combined with our framework, one straightforward way is to have the auxiliary code $c_a$ encode shape details and feed it through the CAPE decoder for per-vertex offsets.
	We found this works reasonably well in feed-forward mode but not the back-propagation.
	We suspect this might because of the inconsistent gradients from different temporal frames, especially when the pose estimation is not perfectly accurate.
	As a result, the compensated geometry is vaguely correct (\eg bump on the head for some hairstyles) but not precise.
	To improve the stability, we propose a shape compensation network (Shape-Comp), in which a GRU takes the auxiliary code $c_a$ as input and predicts a new latent vector for each temporal frame conditioned on the predicted pose.
	The latent vector is then fed into the graph network to predict per-vertex offset, which is similar to the CAPE decoder. We remove the VAE and adversarial loss as they empirically hurt the performance.
	The GRU enables information exchanging across temporal frames, which is critical for robust back-propagation when running applications like motion completion and prediction.

	\subsection{Training Strategy}
% 	\vspace{-0.1in}
	Neural networks with multiple stages are highly non-linear and could easily fall in the local minimum.
	We advocate a stage-wise training strategy to enhance the training stability.
	Specifically, we first train the shape encoder, pose encoder, point feature extractor and motion encoder jointly with the pre-learned linear motion model.
	Once the model converges, we enable the Motion-Comp and Shape-Comp networks for the end-to-end joint training.
	Similar training strategy has been commonly used by other works, \eg BCNet \cite{jiang2020bcnet}, XNect \cite{mehta2019xnect} and Predicting Human Dynamics \cite{zhang2019predicting}.

	\vspace{1mm}
	\noindent \textbf{Loss Functions}
	Since our reconstructions are registered with SMPL topology, we use per-vertex L1-loss with the ground truth mesh as the objective function.
	To further alleviate the ambiguity between body shape and clothing, we add an L2-loss on the shape code $c_s$ with regard to the ground truth.
	During the first training stage, we use the mesh reconstructed with the motion from LMM for supervision.
	In the second stage, we use added loss on both meshes before and after the Shape-Comp network. 
	The detailed formulation of the loss functions can be found in Supp. Mat.
	%-------------------------------------------------------------------------
	
	\begin{figure*}[tb]
		\centering \includegraphics[width=0.95\linewidth]{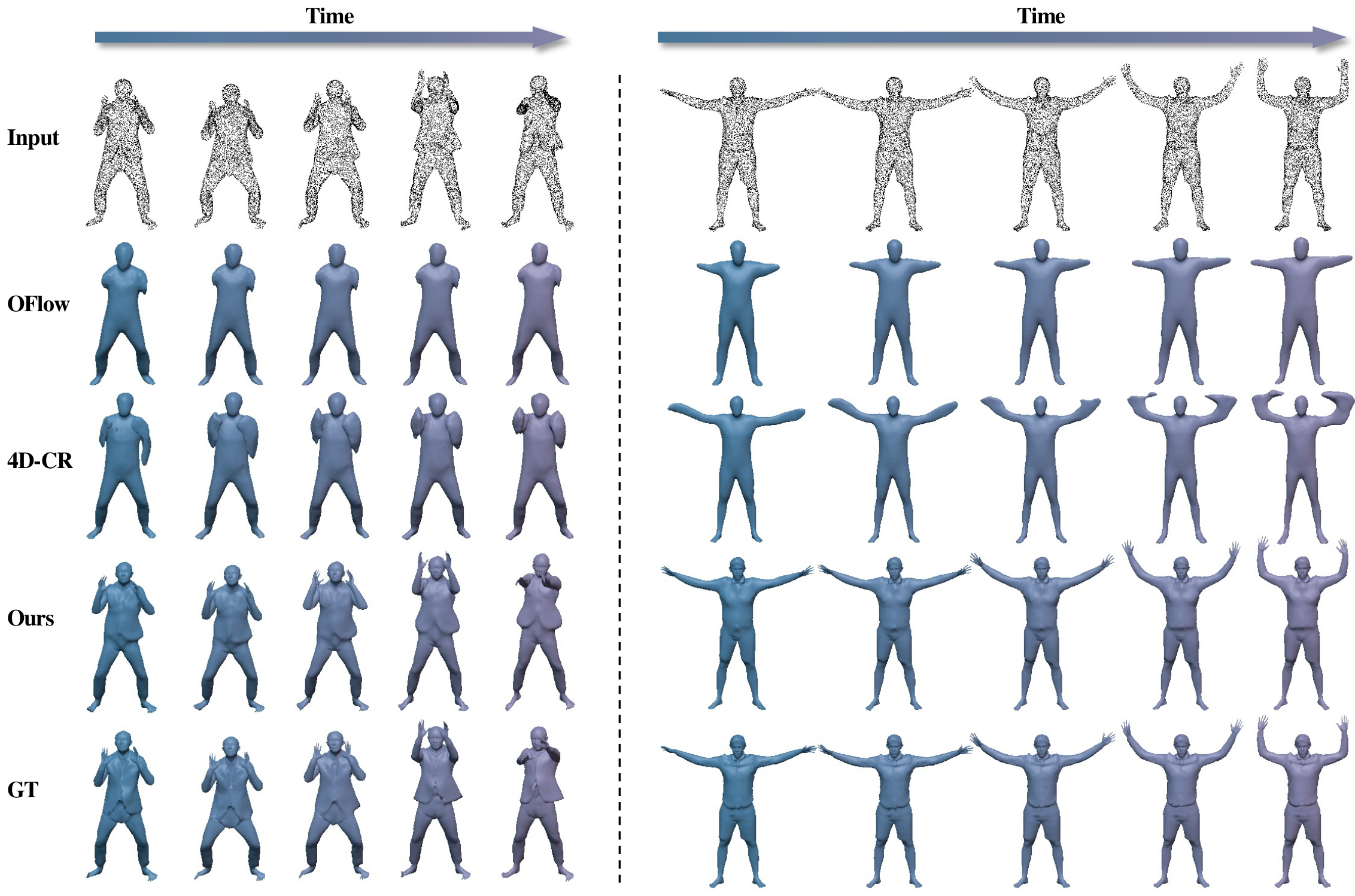}
		\vspace{-0.1in}
		\caption{\textbf{4D Reconstruction.} Given the dense point cloud sequence (Row 1) uniformly sampled from the SMPL registered meshes, our method (Row 4) can reconstruct fine-grained meshes with accurate motion, while baseline methods (Row 2, 3) tend to be overly smooth and often have incomplete geometry, \eg missing hands. We uniformly sample 5 frames (out of 30 frames) for visualization.
			\label{fig:4d_recons} }
		\vspace{-0.1in}
	\end{figure*}

	\begin{table*}[tb]
		\centering
		\small
		\scalebox{0.95}{
		\begin{tabular}{p{2.65cm}p{1.75cm}<{\centering}p{1.25cm}<{\centering}p{1.05cm}<{\centering}p{1.05cm}<{\centering}p{1.75cm}<{\centering}p{1.25cm}<{\centering}p{1.05cm}<{\centering}p{1.05cm}<{\centering}}
			\toprule[1px]
			\multicolumn{9}{c}{I. Comparison with Previous 4D Representation Methods} \\
			\midrule
			\multirow{2}{*}{Methods} & \multicolumn{2}{c}{4D Reconstruction} & \multicolumn{2}{c}{Motion Retargeting} & \multicolumn{2}{c}{Motion Completion} & \multicolumn{2}{c}{Future Prediction} \\
			\cmidrule{2-9}
			& IoU $\uparrow$ & CD $\downarrow$ & IoU $\uparrow$ & CD $\downarrow$ & IoU $\uparrow$ & CD $\downarrow$ & IoU $\uparrow$ & CD $\downarrow$ \\
			\cmidrule{1-9}
			OFlow \cite{niemeyer2019occupancy} & 61.5\% & 0.199 & 30.7\% & 0.470 & 65.8\% & 0.181 & 58.8\% & 0.218 \\
			4D-CR \cite{jiang2021learning} & 62.9\% & 0.165 & 47.3\% & 0.296 & 76.6\% & 0.128 & 64.0\% & 0.200 \\
			Ours & \textbf{73.3\%} & \textbf{0.093} & \textbf{70.7\%} & \textbf{0.100} & \textbf{90.3\%} & \textbf{0.031} & \textbf{71.7\%} & \textbf{0.121} \\
			\toprule[1px]
			\multicolumn{9}{c}{II. Comparison with Human Body Estimation Methods (Forward)} \\
			\midrule
			\multirow{2}{*}{Methods} & \multicolumn{4}{c}{Shape and Motion Recovery} & \multicolumn{4}{c}{Motion Retargeting} \\
			\cmidrule(lr){2-5} \cmidrule(lr){6-9}
			& PA-MPJPE $\downarrow$ & MPJPE $\downarrow$ & PVE $\downarrow$ & Accel $\downarrow$ & PA-MPJPE $\downarrow$ & MPJPE $\downarrow$ & PVE $\downarrow$ & Accel $\downarrow$  \\
			\cmidrule(lr){1-5} \cmidrule(lr){6-9}
			HMMR \cite{kanazawa2019learning} & 87.8 & 102.1 & 89.2 & 20.9 & 85.7 & 98.0 & 86.9 & 19.4 \\
			VIBE \cite{kocabas2020vibe} & 45.3 & 54.3 & 47.6 & 13.4 & 46.3 & 54.1 & 47.0 & 12.8 \\
			4D-CR-SMPL \cite{jiang2021learning} & 59.2 & 68.5 & 59.5 & 9.9 & 62.4 & 73.2 & 63.7 & 10.1 \\
			4D-CR-SMPL$^\ast$ \cite{jiang2021learning} & 49.8 & 57.7 & 49.8 & 8.9 & 52.2 & 59.6 & 51.6 & 8.7 \\
			Ours & \textbf{38.4} & \textbf{44.9} & \textbf{39.2} & \textbf{8.8} & \textbf{39.5} & \textbf{45.2} & \textbf{39.0} & \textbf{8.6} \\
			\toprule[1px]
			\multicolumn{9}{c}{III. Comparison with Human Body Estimation Methods (Backward)} \\
			\midrule
			\multirow{2}{*}{Methods} & \multicolumn{4}{c}{Motion Completion} & \multicolumn{4}{c}{Future Prediction} \\
			\cmidrule(lr){2-5} \cmidrule(lr){6-9}
			& PA-MPJPE $\downarrow$ & MPJPE $\downarrow$ & PVE $\downarrow$ & Accel $\downarrow$ & PA-MPJPE $\downarrow$ & MPJPE $\downarrow$ & PVE $\downarrow$ & Accel $\downarrow$  \\
			\cmidrule(lr){1-5} \cmidrule(lr){6-9}
			HMMR \cite{kanazawa2019learning} & 146.5 & 141.6 & 148.3 & 48.7 & 148.4 & 142.9 & 146.9 & 48.3 \\
			Zhang \etal \cite{zhang2019predicting} & -- & -- & -- & -- & 134.7 & 146.5 & 143.4 & 23.0 \\
			4D-CR-SMPL \cite{jiang2021learning} & 87.3 & 67.3 & 66.9 & 14.1 & 91.9 & 77.9 & 77.1 & 11.3 \\
			Ours & \textbf{53.8} & \textbf{42.7} & \textbf{41.7} & \textbf{9.4} & \textbf{73.1} & \textbf{62.8} & \textbf{59.7} & \textbf{11.2} \\
			\toprule[1px]
		\end{tabular}
		}
		\vspace{-0.1in}
		\caption{\label{tab:result_summary}\textbf{Comparison to SoTA methods on various tasks.} For evaluation, we adopt Volumetric IoU (IoU) and Chamfer Distance (CD) \cite{Occupancy_Networks} for comparisons with free-form methods (block I), and several standard metrics following \cite{kanazawa2019learning,kocabas2020vibe} for SMPL-based methods (block II\&III, the numbers are measured in \textit{mm}). $^\ast$ denotes the input point cloud sequence has temporal correspondence.}
		\vspace{-0.18in}
	\end{table*}
	
% 	\vspace{-0.1in}
	\section{Experiments}
% 	\vspace{-0.1in}
	In this section, we perform extensive experiments to verify the efficacy of our method. First, we evaluate the capacity of our representation for encoding accurate shape and motion on the tasks of 4D reconstruction and human shape and motion recovery.
	We then demonstrate that a large variety of 4D related applications, including motion retargeting, completion, and prediction can be achieved with high quality with our representation.
	Finally, we provide an ablation study to test the impact of each component in our framework on the reconstruction quality.

	\vspace{1mm}
	\noindent \textbf{Dataset} We use the CAPE dataset~\cite{CAPE:CVPR:20,pons2017clothcap} for training and evaluating, which is a dataset of 3D dynamic clothed humans containing 10 male and 5 female subjects wearing different types of outfits.
	More than 600 motion sequences of large pose variations are provided. In each sequence, the clothed body shapes are captured at 60 FPS along with corresponding meshes in the canonical pose and pose parameter of each frame.
	Overall, the dataset provides good diversity on both 3D geometry and motion.
	Following OFlow \cite{niemeyer2019occupancy}, we divide all sequences in CAPE into subsequences of 30 frames.
	We use subsequences from 488 motion sequences for training, and randomly sampled 2000 subsequences from the other 123 motion sequences for testing.
	
	\noindent \textbf{Implementation} We use PyTorch to implement the model, and train with the Adam optimizer \cite{kingma2014adam}. In the first stage, the learning rate is $10^{-4}$ with batch size 16. In the second stage, the initial learning rate is set to $10^{-4}$ and dropped to $10^{-5}$ after 200K iterations with batch size 4 due to the limitation of GPU memory. We use 4 NVIDIA GeForce RTX 2080Ti GPU cards.
		
	\noindent \textbf{Evaluation} 
	To measure the difference between the prediction and ground truth 3D shape, we use Chamfer Distance (CD) and Volumetric IoU (IoU) \cite{Occupancy_Networks} for free-form geometry and Per Vertex Error (PVE) for SMPL registered shape.
	To measure the accuracy of motion, we use Procrustes-aligned mean per joint position error (PA-MPJPE), mean per joint position error (MPJPE), and acceleration error ($mm/s^2$) computed on 45 keypoints which include 24 joints and 21 keypoints on the face, feet and hands.
	Please refer to \cite{Occupancy_Networks,kanazawa2019learning,kocabas2020vibe} for more details about these metrics. For temporal sequences, we take the mean score of all frames.

	\subsection{Representation Capability}
	\label{sec:motion_rec}
    % We first show our representation is capable to encode and reconstruct 4D human with accurate motion and geometry.
    We first show that our representation is capable of encoding and reconstructing human sequences with correct motion and geometry.
    
    % \vspace{1mm}
    \noindent \textbf{4D Reconstruction}
	We compare to state-of-the-art 4D representation, Occupancy Flow (OFlow)\cite{niemeyer2019occupancy} and 4D-CR \cite{jiang2021learning}, on mesh reconstruction from sampled point cloud inputs.
	As shown in Tab. \ref{tab:result_summary} (I), our method significantly outperforms other methods on the 4D reconstruction accuracy.
	Qualitative results are shown in Fig. \ref{fig:4d_recons} for two temporal sequences.
	OFlow tends to produce incomplete geometry with missing hands, and the results from 4D-CR are overly smoothed, \eg around face and hands.
	On the contrary, thanks to the human prior, our results are significantly better than others with complete geometry, correct motion, and rich details such as fingers, clothes, and hair.
	It is also worth noting that both OFlow and 4D-CR require point clouds with temporal correspondence as input, whereas our method can take unregistered point cloud sequences which is more convenient for many applications.
	
% 	\vspace{-1mm}
	\noindent\textbf{Shape and Motion Recovery} \label{sec:motion_recovery}
	We then study the performance of our motion model, which consists of a linear motion model and per-frame compensations recovered from the auxiliary code.
	As baselines, we compare to SoTA video-based human shape and pose estimation methods HMMR \cite{kanazawa2019learning} and VIBE \cite{kocabas2020vibe}.
	Originally designed for color images input, we replace their image encoder with the point cloud encoder in our setup.
	Moreover, as an additional baseline, we extend 4D-CR \cite{jiang2021learning} to an SMPL-based version by replacing their implicit occupancy decoder with the SMPL decoder so that it also benefits from the human prior.
	Since all of these methods produce only unclothed SMPL defined shapes, we disable our Shape-Comp network and use the output from the SMPL decoder for fair comparisons.
	All the baseline methods are retrained on our dataset.
	For extending 4D-CR, we train two models with registered point clouds as in their work (4D-CR-SMPL$^\ast$) and unregistered point clouds like us (4D-CR-SMPL) respectively.
	The quantitative comparisons are shown in Tab. \ref{tab:result_summary} (II).
	Our method achieves more accurate motion estimation (as measured at body keypoints by PA-MPJPE, MPJPE, acceleration error) and SMPL shape (as measured by PVE) than HMMR and VIBE.
	4D-CR-SMPL performs relatively poorly when the input point cloud is unordered and gets much better once given tracked point clouds (4D-CR-SMPL$^\ast$) but still performs worse than our method. We provide additional comparisons to the recent motion-based human body estimation method HuMoR \cite{rempe2021humor} with the task of shape and motion recovery from point clouds via auto-decoding in Supp. Material.
	
% 	\vspace{-0.1mm}
	\subsection{Applications}
	\vspace{-1mm}
    Our representation can support various applications.
	Note that for all applications, the encoder and decoder are both fixed after training.
	
	\noindent \textbf{Motion Retargeting} The goal of motion retargeting is to transfer the motion sequence from one subject to another.
	Traditional methods typically require manual works, \eg provide correspondences between source and target identities \cite{sumner2004deformation}, to fulfill such a task.
	
	We achieve motion retargeting without any human intervention.
	Taking two point cloud sequences, one as the identity ($I$) and the other as the motion ($M$), we feed both into our compositional encoder to get SMPL parameters and latent codes for each $(c_s^I, c_p^I, c_m^I, c_a^I)$ and $(c_s^M, c_p^M, c_m^M, c_a^M)$.
	We then conduct the motion retargeting by using $(c_s^I, c_p^M, c_m^M)$ for linear motion model, $c_a^M$ for Motion-Comp network, and $c_a^I$ for Shape-Comp network.
	Note that two $c_a$ are used for Motion-Comp and Shape-Comp networks separately as they encode motion and shape information respectively.
	
\begin{figure}[t]
		\centering \includegraphics[width=0.9\linewidth]{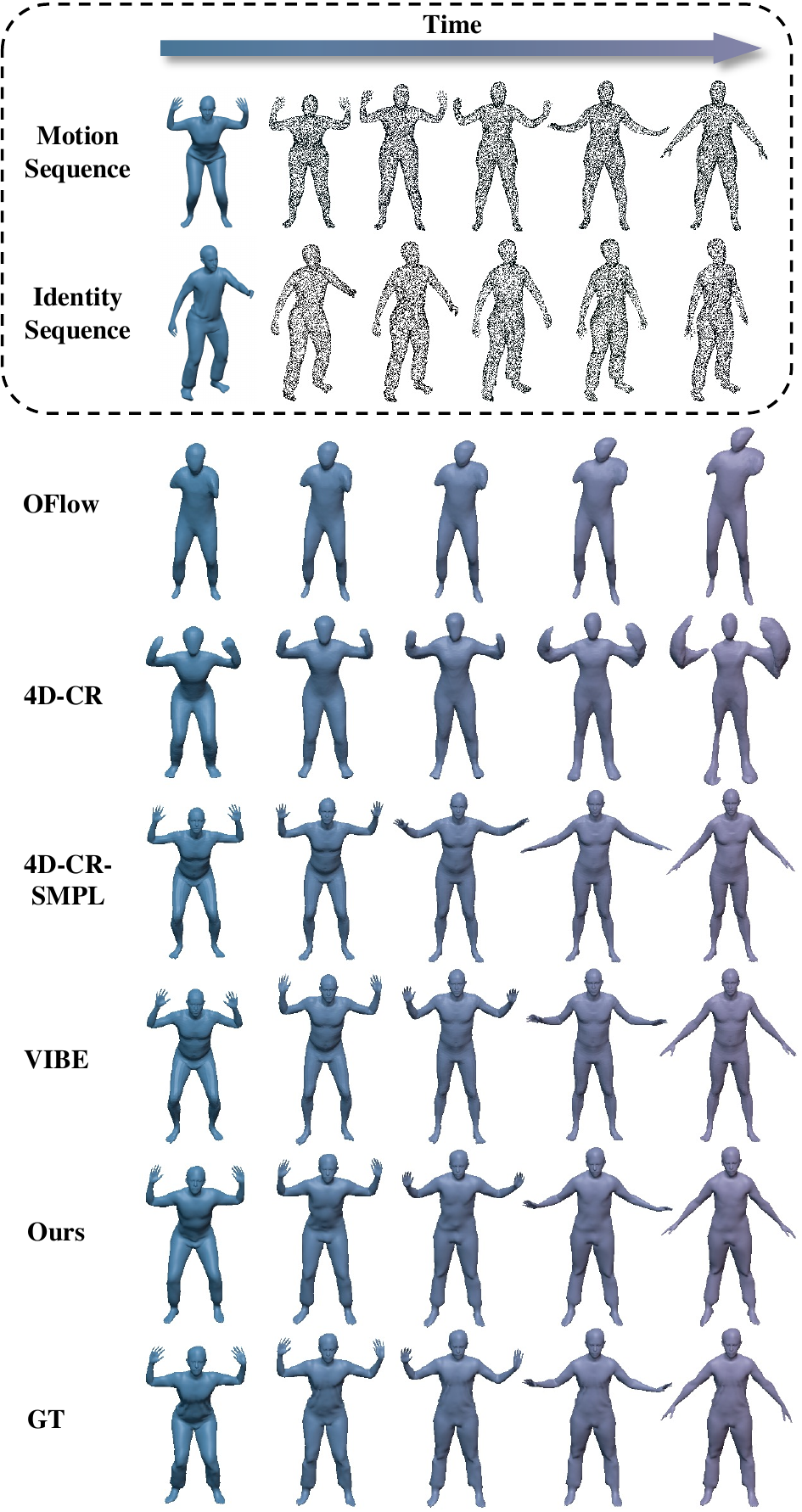}
		\vspace{-0.1in}
		\caption{\textbf{Motion Retargeting.} Our goal is to transfer the human movements of the motion sequence (Row 1) to the people in the identity sequence (Row 2). We can accurately transfer the motion to the new identity and keep the original geometry details, \eg garments and hairstyle, at the same time (Row 7). The free-form baselines (Row 3, 4) either fail due to shape and motion entanglement or produce more artifacts over time due to error accumulation. The SMPL-based baselines (Row 5, 6) also fulfill retargeting but are inaccurate, and they can only represent unclothed body.
			\label{fig:motion_transfer}}
		\vspace{-0.25in}
	\end{figure}

	For evaluation purpose, we randomly sampled 100 pairs of identity and motion sequences with $L=30$ frames.
	We use the provided ground truth shape in the canonical pose and pose parameters provided by CAPE \cite{CAPE:CVPR:20} dataset to generate motion retargeted ground truth sequences.
	We compare with free-form geometry based approach (OFlow and 4D-CR) on the full geometry (in Tab.~\ref{tab:result_summary} (I)) and SMPL based approach (HMMR, VIBE, 4D-CR-SMPL) on the unclothed body mesh from SMPL (in Tab.~\ref{tab:result_summary} (II)).
	Our method significantly outperforms OFlow and 4D-CR.
	As shown in a qualitative example in Fig. \ref{fig:motion_transfer}, our method produces much more complete motion retargeting results.
	Note how clothing details are successfully transferred, \eg long trousers in the identity sequence compared to shorts in the motion sequence.
	Our method also outperforms all the human body estimation methods, showing that the compositional encoder is more effective in extracting correct information from inputs and facilitating motion recovering.
	
	\vspace{1mm}
	\noindent \textbf{Motion Completion}
	Our representation can also fulfill fitting tasks in the auto-decoding fashion, in which the SMPL parameters and latent codes are optimized to produce output similar to the observation.
	With this, our representation can perform motion completion, where the goal is to predict the missing data in a dynamic human sequence.
	For evaluation, we randomly choose 100 sequences with 30 frames from our test set.
	For each sequence, we randomly pick 15 frames as the observation, optimize the SMPL parameters and latent codes, reconstruct the full sequence, and then measure the geometry accuracy on the other 15 frames.
	Note that we use Chamfer loss with the additional prior terms borrowed from IPNet \cite{bhatnagar2020ipnet} on uniformly sampled points instead of PVE to simulate the case in real applications, where the observed meshes may not be registered.

	Comparisons to free-form based methods and SMPL based methods are shown in Tab. \ref{tab:result_summary} (I) and (III) respectively, and the qualitative results are in Supp. Material.
	Zhang \etal \cite{zhang2019predicting} uses a similar motion model to HMMR \cite{kanazawa2019learning}, so we only evaluate one of them.
	Overall, our method consistently outperforms all the other methods. 
	
	Moreover, we compare the robustness of auto-decoding based fitting using our Shape-Comp network against the naive CAPE decoder, and show the error of completion w.r.t the amount of random noise added to the observed frames in Fig. \ref{fig:ablation_curve} (b).
	The error of our model is consistently lower than the naive CAPE decoder and deteriorates less with increasing noise.
	This is presumably because CAPE performs per-frame optimization, which may confuse the latent space if the gradients are not consistent from temporal frames, whereas we use GRU to model the temporal sequence for more robustness.
	
	Last but not least, our model can also complete the temporal sequences from partial spatial observation.
	To show this, we generate one depth image per-frame from a camera rotating concurrently with the motion of the 3D shape, and run auto-decoding based fitting to complete the sequence.
	This can be also considered as a typical non-rigid fusion with known camera poses.
	We show the qualitative and quantitative comparisons to NPMs \cite{palafox2021npms} in Supp. Material.
	
\begin{figure}[htb]
		\centering
		\vspace{-0.15in}
		\includegraphics[width=0.9\linewidth]{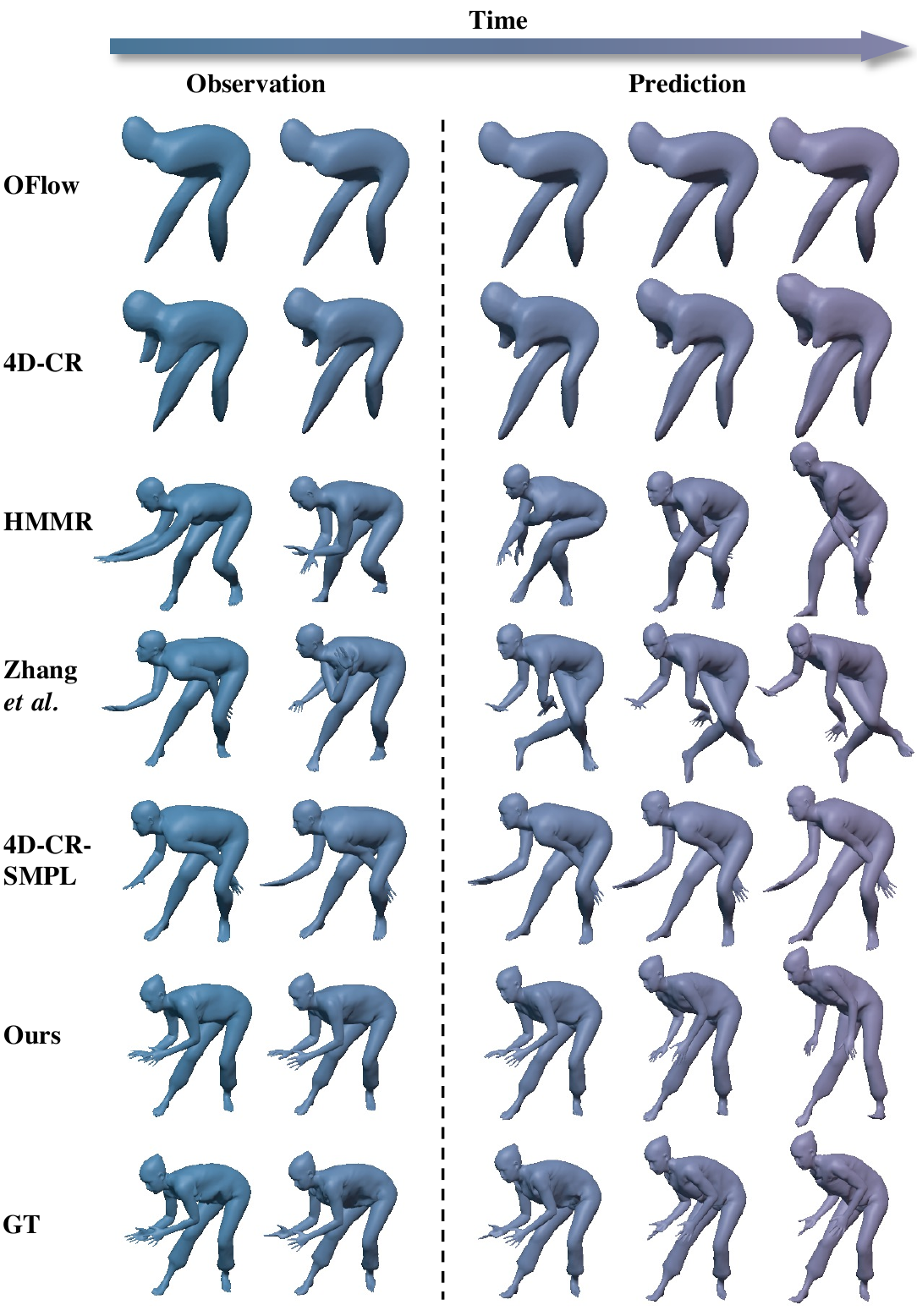}
		\vspace{-0.1in}
		\caption{\textbf{Future Prediction.} We extrapolate 10 future temporal frames based on 20 past observed frames. The baseline methods (Row 1-5) either produce unsatisfactory geometry or stuck into the unnatural pose, while our method (Row 6) successfully keeps the movement trend and produces reasonable prediction of future motion.
		The meshes on the left and right are reconstructions of the observations and predictions for the future time steps, respectively.
		\label{fig:future_pred} }
		\vspace{-0.25in}
	\end{figure}

	\vspace{1mm}
	\noindent \textbf{Future Prediction}
	Our representation also supports future prediction.
	Specifically, we run a fitting algorithm on the first 20 frames, generate the SMPL parameters and latent codes, and then reconstruct the full sequence to predict the 10 frames in the future.
	Tab.~\ref{tab:result_summary} (I, III) and Fig. \ref{fig:future_pred} show the comparison to the previous methods.
	Again, we obtain significantly better performance than other 4D representation methods (OFlow and 4D-CR).
	When comparing only the motion accuracy using SMPL mesh with previous work on motion prediction (HMMR \cite{kanazawa2019learning}, Zhang \etal \cite{zhang2019predicting} and 4D-CR-SMPL \cite{jiang2021learning}), our method still achieves better performance. Moreover, we empirically found that, though given pose prior terms during backward optimization, these baseline methods are more easily to produce unnatural poses than us, and predict unreasonable motions as shown in Fig. \ref{fig:future_pred}, possibly because our PCA-based motion model provides regularization and global context for output motions.
	
	\subsection{Ablation Study}
% 	\vspace{-0.05in}
	In this section, we perform an ablation study and show the quantitative results to demonstrate the effect of the major designs in our method. The visualization example can be found in Supp. Material.

\begin{figure}[tb]
		\hspace{-0.1in}
		\subcaptionbox{Ablation Study}{
			\vspace{0.1in}
			\begin{minipage}{0.45\linewidth}
				% \centering
				%			\parbox[][2.7cm][c]{\linewidth}{
				\scalebox{0.6}{
					\renewcommand\tabcolsep{2pt}
					\begin{tabular}{lcccc}
						
						\toprule[1px]
						& PA-MPJPE $\downarrow$ & MPJPE $\downarrow$ & PVE $\downarrow$ & Accel $\downarrow$ \\
						\midrule
						\textbf{-} GRU Enc. & 49.6 & 57.0 & 49.6 & 10.6 \\
						\midrule
						\textbf{-} LMM & 40.2 & 46.8 & 41.2 & \textbf{8.8} \\
						\textbf{-} Motion $c_a$ & 43.7 & 50.4 & 43.4 & 9.0 \\
						Full Model & \textbf{38.4} & \textbf{44.9} & \textbf{39.2} & \textbf{8.8} \\
						\midrule
						\textbf{-} Shape $c_a$ $^{\dag}$ & -- & -- & 43.8 & -- \\
						Full Model$^{\dag}$ & -- & -- & \textbf{42.0} & -- \\
						\toprule[1px]					
					\end{tabular}
				}	
			\end{minipage}
		}
		\hfill
		\subcaptionbox{Noise Tolerance}{
			\label{fig:noise_curve}
			\begin{minipage}{0.4\linewidth}
				%			\centering
				\includegraphics[width=1\linewidth]{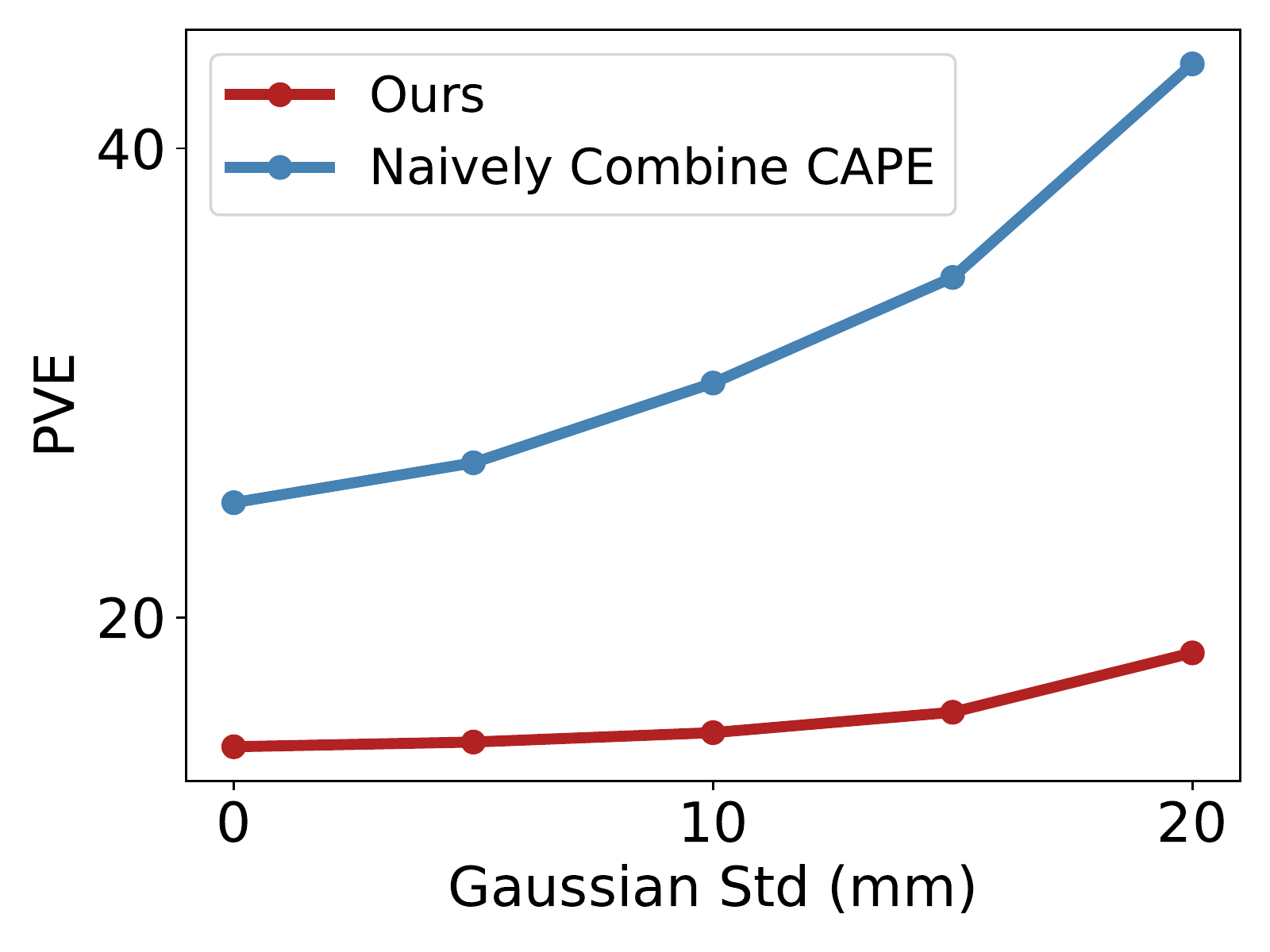}
			\end{minipage}
		}
		\vspace{-0.1in}
		\caption{(a) \textbf{Ablation study.} 
			We verify the effectiveness of our temporal encoder by replacing the GRU with the modified PointNet used in \cite{niemeyer2019occupancy,jiang2021learning} (Row 1). Additionally, we remove the major modules in our framework in turn to demonstrate the effect of different components (Row 2, 3, 5).
			$^{\dag}$ denotes we compute the metrics with the ground truth clothed mesh. (b) \textbf{Robustness against noise.} The x-axis is the standard deviation of added Gaussian noise, and the y-axis is Per Vetex Error (PVE, lower is better). }
		\label{fig:ablation_curve}
		\vspace{-0.2in}
	\end{figure}

	\noindent \textbf{Motion Model} 
	We first study the effect of the linear motion model and auxiliary code for motion recovery.
	We compare the ablation cases on the output of the SMPL decoder with the registered SMPL model on the ground truth mesh, which removes the free-form deformation and allows us to focus on the motion quality.
	In Fig. \ref{fig:ablation_curve} (a), we show the performance of our model removing the linear motion model (``-LMM''), which directly updates the initial pose code, or removing Motion-Comp network (``-Motion $c_a$''), which only relies on the linear motion model (LMM) for motion recovery. In either case, the motion accuracy drops consistently as measured by all metrics, indicating the necessity of combining the prior model with learned compensations.

	\noindent \textbf{Shape Model} 
	We then verify if the auxiliary code helps to recover the detailed geometry.
	In Fig.~\ref{fig:ablation_curve} (a), we show the performance of the final mesh without and with auxiliary code-driven shape compensation (the last two rows), which is measured by PVE between the output mesh with the ground truth clothed mesh.
	The advantage of the shape compensation can also be found in Fig.~\ref{fig:motion_transfer} and \ref{fig:future_pred}, which show that the auxiliary code helps to improve the geometry details when comparing our results with the SMPL outputs from VIBE, HMMR or 4D-CR-SMPL.

	\noindent \textbf{Encoder}
	Last but not least, we verify the effectiveness of our GRU-based temporal encoder.
	We replace our temporal encoder with the PointNet adopted in OFlow \cite{niemeyer2019occupancy} and 4D-CR \cite{jiang2021learning}, and see a significant performance drop (``-GRU Enc.''), which shows our GRU-based encoder helps to extract temporal information from the point cloud sequences without temporal correspondence.

	%-------------------------------------------------------------------------
	\section{Conclusion}
	\vspace{-1mm}
	This paper introduces H4D, a compact and compositional neural representation for 4D human captures, which combines the merits of both the prior-based and free-form solutions. A novel framework is designed for learning our representation, which encodes the input point cloud sequences into the SMPL parameters of shape and initial pose, and latent codes of motion and auxiliary information. Extensive experiments on 4D reconstruction, shape and motion recovery, motion retargeting, completion and prediction validate the efficacy of the proposed approach.

	\section*{Acknowledgements}
	\vspace{-1mm}
	This work was supported in part by NSFC Project (62176061),  Shanghai Municipal Science and Technology Major Project (2018SHZDZX01). The corresponding authors are Xiangyang Xue, and Yanwei Fu.

	%-------------------------------------------------------------------------

	%%%%%%%%% REFERENCES
	{\small
		\bibliographystyle{ieee_fullname}
		\bibliography{egbib}
	}

    % Supplementary Material
    \clearpage
    \noindent \textbf{\Large Supplementary Material}
    \vspace{0.1in}
    \setcounter{section}{0}
    
    %%%%%%%%% BODY TEXT
In this supplementary material, we provide implementation details, results about the generalization ability of our motion model, extended comparisons to related works, visualization of principal components, additional qualitative results, run-time comparison, and discussions about limitations, future work and broader impact of our approach.

\begin{figure*}[htb]
\centering 

\subfloat[Spatial Encoder]{                  
    \label{fig:spatial_enc}            
\includegraphics[width=0.52\linewidth]{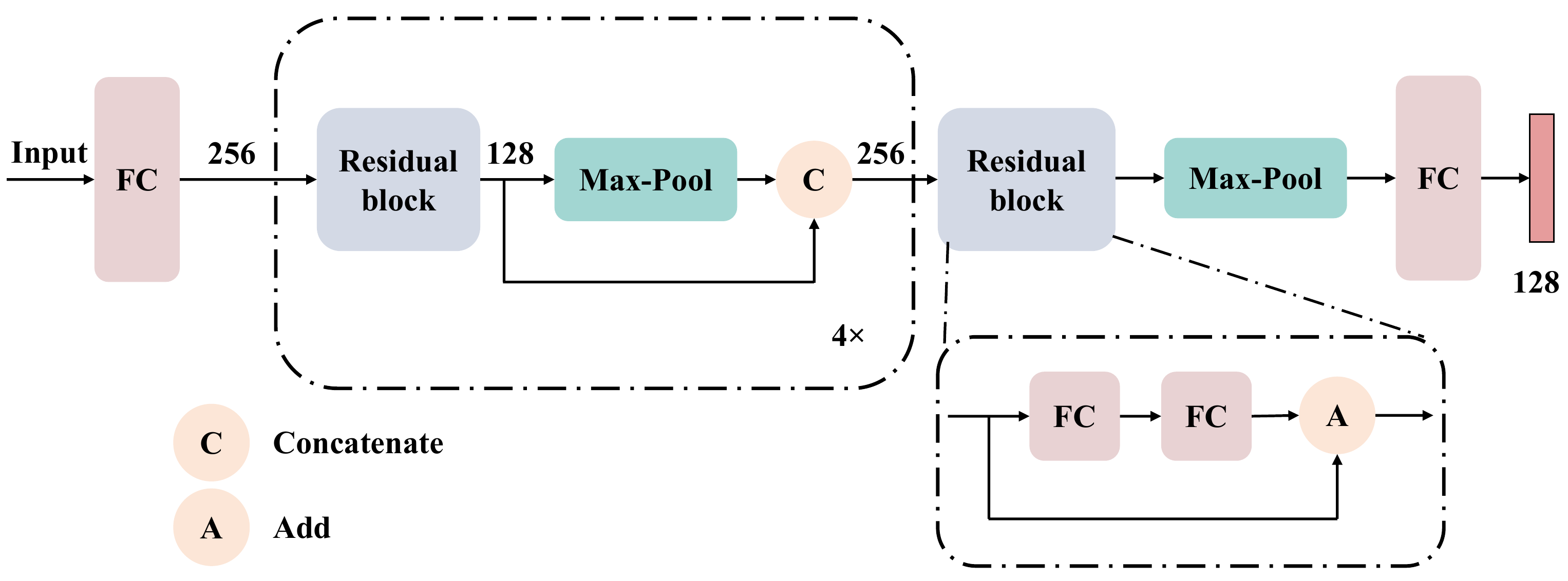}
}
\subfloat[Temporal Encoder]{                    
    \label{fig:temporal_enc}            
\includegraphics[width=0.4\linewidth]{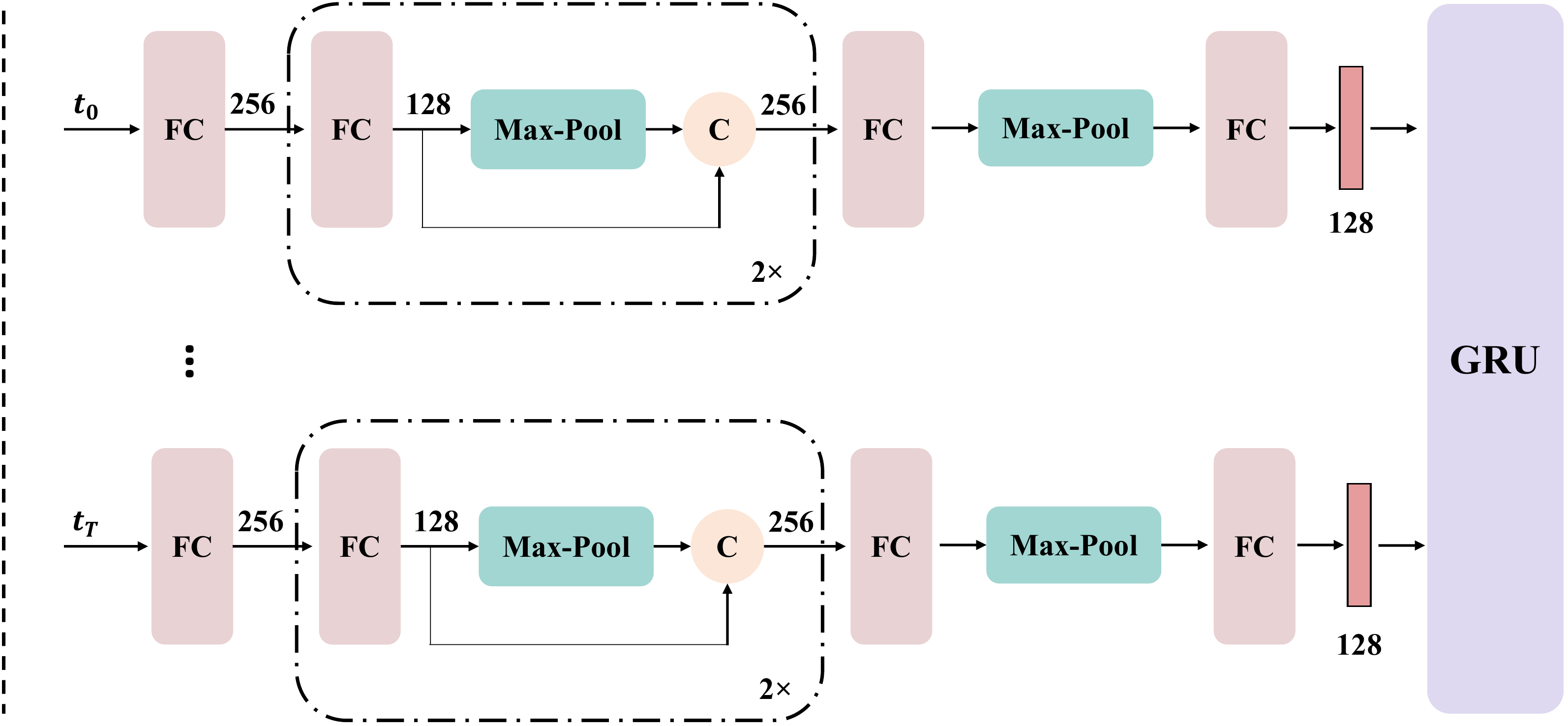}
}
\vspace{0.1in}
\subfloat[Motion-Comp Network]{       \centering    
\label{fig:motion_comp}           
\includegraphics[width=0.45\linewidth]{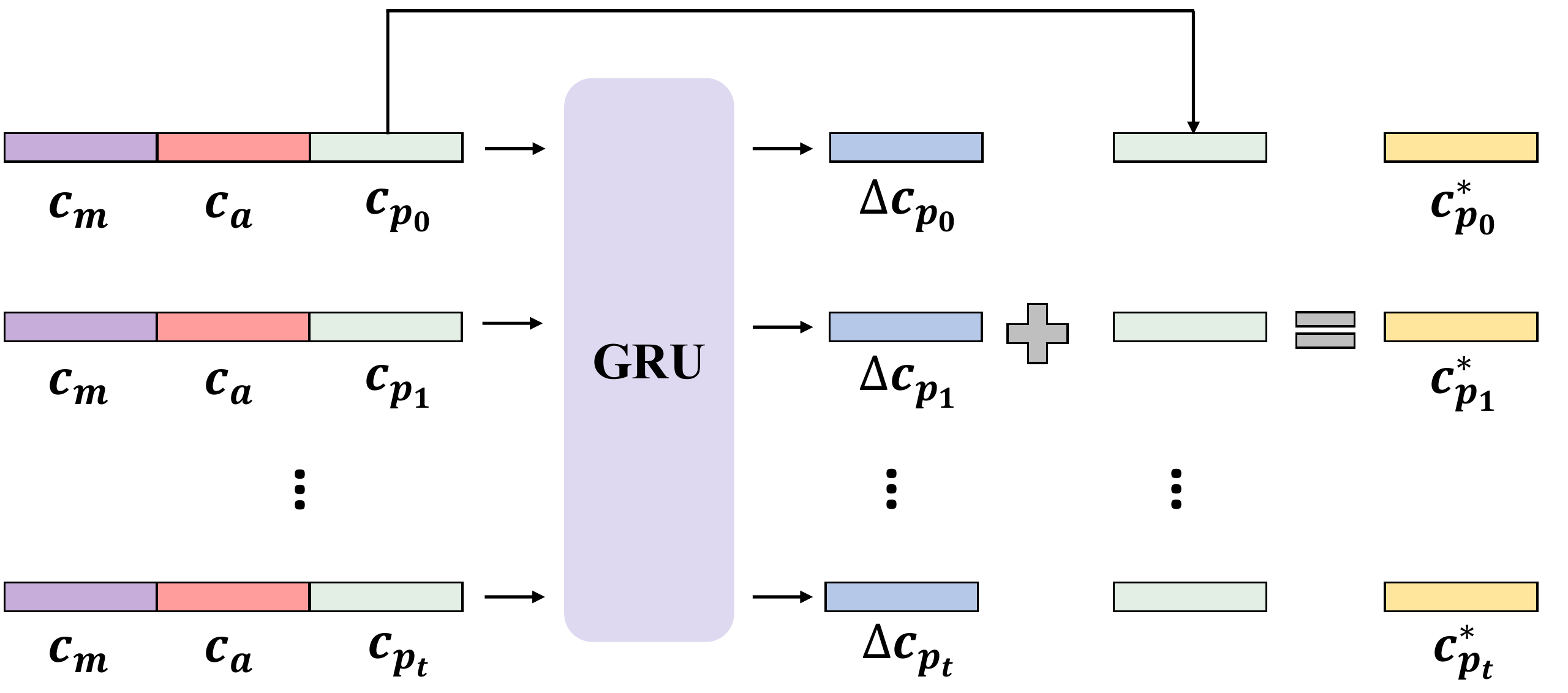}
}
\subfloat[Shape-Comp Network]{       \centering    
\label{fig:shape_comp} 
\includegraphics[width=0.42\linewidth]{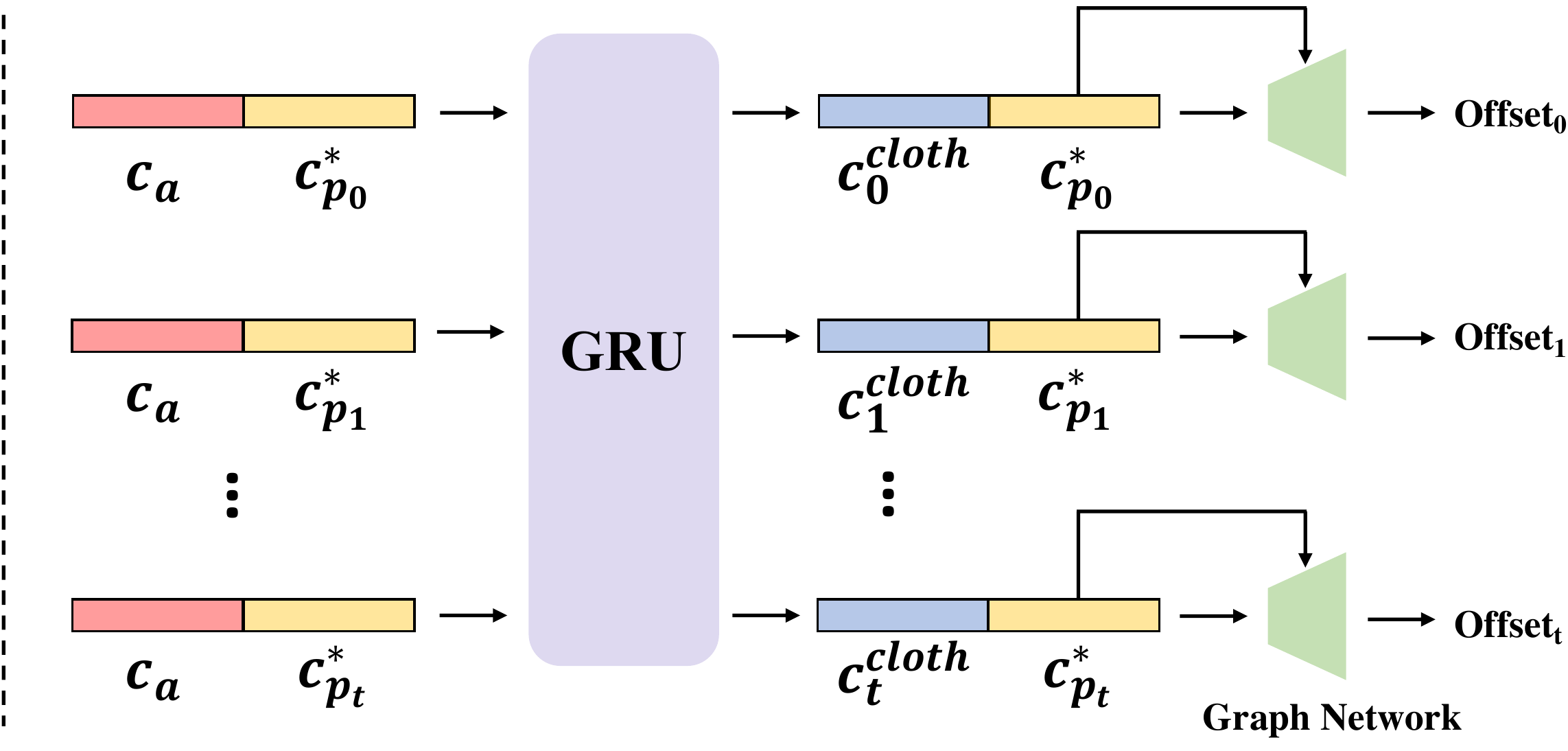}
}

\caption{\textbf{Detailed network architectures in our framework.}}
\vspace{-0.15in}
\label{fig:architecture} 
\end{figure*}

\section{Implementation Details}
In this section, we first provide network architectures used for the compositional encoder, Motion-Comp and Shape-Comp networks in our framework. Next, we explain the strategy of choosing the number of principal components for our linear motion model. Finally, we discuss more details in our experiments.

\subsection{Network Architecture}
\noindent \textbf{Compositional Encoder} Both the shape encoder and the initial pose encoder take point cloud of the first time step as input, and we adopted the same architecture as the spatial encoder in Occupancy Flow (OFlow) \cite{niemeyer2019occupancy}.
The network is a variation of PointNet \cite{qi2016pointnet} which has five residual blocks as show in Fig. \ref{fig:spatial_enc}. Each of the first four blocks has an additional max-pooling operation to obtain the aggregated feature of size $\left(B,1,C\right)$ where $C$ denotes the dimension of hidden layers, and an expansion operation (repeat the pooled feature to the size $\left(B,N,C\right)$) to make it suitable for concatenation.
The output of the fifth block is passed through a max-pooling layer and a fully connected layer to get the final outputs of dimension 10 for shape parameter and 72 for initial pose parameter. 

Our temporal encoder, for the purpose of learning motion and auxiliary codes, is composed of a point feature extractor (shallow PointNet) and a double layers GRU \cite{cho2014learning}, as shown in Fig. \ref{fig:temporal_enc}. The shallow PointNet extracts spatial features for each input point cloud, which has 3 hidden layers with hidden sizes equal to 128. We use the same max-pooling and concatenating operations as the spatial encoder. Then the per-frame features are processed sequentially by the GRU layer to provide the latent vector of dimension 90 for motion code and 128 for auxiliary code.

\vspace{1mm}
\noindent \textbf{Motion-Comp Network} We design a conditional GRU for our Motion-Comp network to learn the compensation of the input motion sequence. Specifically, we use the motion code $c_m$ and auxiliary code $c_a$ as conditions, copy and concatenate them with the pose parameter of each time step estimated by our linear motion model.
The detailed architecture is shown in Fig. \ref{fig:motion_comp}. The output of the conditional GRU is the motion compensation, and we apply a residual connection to obtain the refined motion sequence, \ie per-frame poses. We can recover body mesh sequences with the predicted shape and per-frame pose codes by using SMPL decoder, here we use the neutral shape model as in previous work \cite{kanazawa2018end,kolotouros2019learning,kocabas2020vibe}.

\vspace{1mm}
\noindent \textbf{Shape-Comp Network}
We propose a Shape-Comp network, in which a conditional GRU takes the auxiliary code $c_a$ as input and predicts a new latent vector for each temporal frame conditioned on the predicted pose (we follow CAPE to represent each joint with the flattened rotational matrix and filter the joints that are not related to clothing). The latent vector of each frame is then fed into the graph network to predict per-vertex offsets, which is similar to the CAPE decoder. We remove the one-hot vector of clothing type and only use the predicted pose as condition since we do not focus on the generative task. The architecture is shown in Fig. \ref{fig:shape_comp}.

\vspace{1mm}
\noindent \textbf{Implementation of GRUs} We use the standard API of GRU provided by PyTorch.
All the GRUs in our framework share the same architecture, which has 2 layers with the hidden size of 512, except we apply an additional linear layer for each GRU to transform the output dimensions for different modules.

\subsection{Linear Motion Model}
For the linear motion model (Section 3.2 in the main paper), we employ the
% most classical 
Principal Component Analysis (PCA) to model the per-frame difference of the pose parameter regarding the first frame in a sequence. As stated in Sec. 3.2 of the main paper, we run PCA separately for the global orientation (\ie pelvis) and the remaining body joint rotations. Inspired by Urtasun \etal \cite{urtasun2006temporal}, we choose the number of PCA components depending on the fraction of the total variance of the training data that is captured by the subspace, denoted by $Q(m)$:
\begin{equation}
Q(m)=\frac{\sum_{i=1}^{m} \lambda_{i}}{\sum_{i=1}^{M} \lambda_{i}}
\end{equation}
where $m$ controls the number of principal components, $\lambda_i$ are ordered eigenvalues of the data covariance matrix such that $\lambda_i\geq\lambda_{i+1}$, and $M$ is the total number of eigenvalues. In our experiments, we choose $m=4$ for the global rotation and $m=86$ for the remaining body joints rotation, which satisfy $Q(m)>0.9$. We visualize some principal components in Fig. \ref{fig:pca1} and \ref{fig:pca2} (Sec. \ref{sec:vis_pca}).

\subsection{Experiment Details}\label{sec:experiment_detail}
\noindent \textbf{Loss Functions}
Given an input point cloud sequence, our model generates one shape code $c_s$ and three mesh sequences $\mathbf{X}_{\text{linear}}$, $\mathbf{X}_{\text{motion}}$ and $\mathbf{X}_{\text{shape}}$, which correspond to the outputs of LMM, Motion-Comp network and Shape-Comp network respectively. Each sequence has $L=30$ mesh frames and each mesh has $K=6890$ vertices. 
We also have the ground truth shape parameter $c_s^{\ast}$ and posed SMPL body mesh sequence $\mathbf{Y}_{\text{body}}$. Furthermore, we compute ground truth offsets sequence by $\mathbf{Y}_{\text{offset}}=\mathcal{M}_{\text {clothed}}-\mathcal{M}_{\text {SMPL}}$, where $\mathcal{M}_{\text {clothed}}$ and $\mathcal{M}_{\text {SMPL}}$ stand for the vertices of the clothed human mesh and corresponding SMPL body mesh in the canonical pose respectively.

Then we define the reconstruction loss as the per-vertex $L_1$ error with the ground truth mesh
\begin{equation}
\mathcal{L}_{\text{r}}\left(\mathbf{X},\mathbf{Y}\right)=\frac{1}{L K} \sum_{l=1}^{L} \sum_{k=1}^{K} \left\|\mathbf{X}_{l, k}-\mathbf{Y}_{l, k}\right\|_{1}.
\end{equation}

Furthermore, we apply $L_2$ penalization on the predicted shape code to further alleviate the ambiguity between body shape and clothing, given by
\begin{equation}
\mathcal{L}_{\text{s}}\left(\mathbf{c},\mathbf{c^{\ast}}\right)=\left\|\mathbf{c}-\mathbf{c^{\ast}}\right\|_{2}^{2}.
\end{equation}

Finally, the total loss for training can be formulated as
\begin{equation}
\begin{aligned}
\mathcal{L}=\lambda_{s}\mathcal{L}_{\text{s}}\left(\mathbf{c_s},\mathbf{c_s^{\ast}}\right)&+\lambda_{r_1}\mathcal{L}_{\text{r}}\left(\mathbf{X}_{\text{linear}},\mathbf{Y}_{\text{body}}\right)\\ &
+\lambda_{r_2}\mathcal{L}_{\text{r}}\left(\mathbf{X}_{\text{motion}},\mathbf{Y}_{\text{body}}\right)\\ & +\lambda_{r_3}\mathcal{L}_{\text{r}}\left(\mathbf{X}_{\text{shape}},\mathbf{Y}_{\text{offset}}\right),
 \end{aligned}
\end{equation}
we set $\lambda_{s}=\lambda_{r_1}=1$, $\lambda_{r_2}=\lambda_{r_3}=0$ for the first training stage and $\lambda_{s}=\lambda_{r_2}=1$, $\lambda_{r_3}=30$ and $\lambda_{r_1}=0$ for the second stage.

\vspace{1mm}
\noindent \textbf{Backward Experiments}
For our auto-decoding based experiments, \ie completion and prediction, we use the trained model to perform a backward fitting algorithm. Specifically, we remove the encoder, freeze the parameters of the remaining modules and optimize the SMPL parameters and latent codes with back-propagation to produce the outputs as similar to the observations as possible. 

We initialize the SMPL parameters and latent codes with the random vector sampled from a Gaussian distribution $N(0,0.01)$ and use the Adam optimizer \cite{kingma2014adam} with learning rate $3e^{-2}$ to perform back-propagation for 500 iterations. In each iteration, we uniformly sample 8192 points on the surface of the predicted meshes, and compute Chamfer loss \cite{Occupancy_Networks,ravi2020pytorch3d} w.r.t the observed points for penalizing. Additionally, we follow IPNet \cite{bhatnagar2020ipnet} to add pose and shape prior terms, which penalize unnatural output bodies during optimization.

\vspace{1mm}
\noindent \textbf{Completion}
We conduct two different types of motion completion experiments, \ie temporal completion and spatial completion. 
Given a temporal sequence of $L=30$ frames, for temporal completion, we randomly select 15 frames as observation and optimize the SMPL parameters and latent codes to complete the missing frames. We choose HMMR \cite{kanazawa2019learning} and 4D-CR-SMPL (an extension we implement for 4D-CR \cite{jiang2021learning}) as baselines.
To implement 4D-CR-SMPL, we replace their implicit decoder with the SMPL decoder, and set the dimensions of their identity code and initial pose code to 10 and 72 respectively. Then we can obtain the pose code for each time step with the Neural ODE conditioned on the motion code, and input it to the SMPL decoder with the identity code to produce the reconstructed mesh frame.
For 4D-CR-SMPL, we use the model trained on our dataset, and for HMMR, we use the official pretrained model.

The goal of spatial completion is to complete the temporal sequence with partial spatial observation. To this end, we use the raw scanned mesh sequences of CAPE \cite{CAPE:CVPR:20}, and render the depth images of resolution $512\times512$ with the approach illustrated in the main paper (Sec. 4.2) to simulate the real world scenario. We assume the camera poses are known and back project the depth images to obtain partial point clouds. We initialize our codes with the random vectors sampled from a Gaussian distribution $N(0, 0.01)$ with no requirement for an additional initialization step like NPMs \cite{palafox2021npms}, and use the Adam optimizer with learning rate $3e^{-2}$ to perform back-propagation for 500 iterations. Note that in this experiment, we adopt one-directional point-to-surface loss instead of two-directional Chamfer loss due to the partial geometry.
We show some qualitative examples in Fig. \ref{fig:spatial_comp}.

\begin{figure*}[tb]
\centering
\includegraphics[width=1\linewidth]{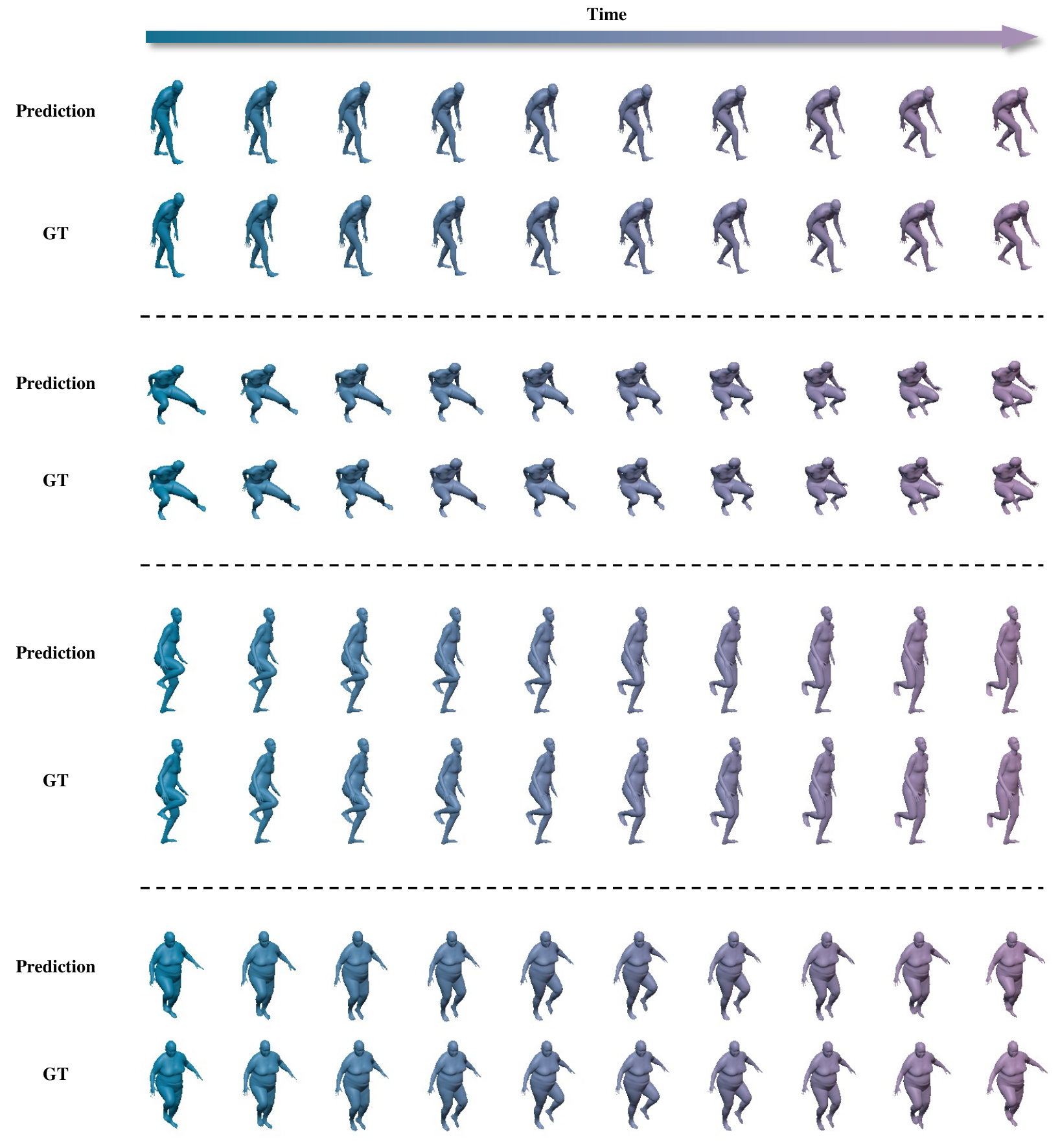}
\caption{\textbf{Results of the novel motions from AMASS dataset.} To investigate the generalization ability of our method, we choose 4 motion sequences from the AMASS dataset, and use our model trained on the CAPE dataset to fit them by using the backward algorithm.}
\label{fig:amass}
\end{figure*}

\section{Generalization of Our Motion Model}
In this section, our goal is to investigate the capacity of our motion model for representing novel motions from another dataset. To this end, we choose some motion sequences from AMASS \cite{mahmood2019amass}, a large 3D MoCap dataset. And then we use our model trained on the CAPE dataset \cite{CAPE:CVPR:20}, perform the similar backward algorithm in completion and prediction experiments to fit the whole sequence of $L=30$. Instead of dense SMPL, we randomly sample 8192 points from the SMPL mesh of each frame as observations, then use the Chamfer loss to the points sampled from the predicted mesh. Prior terms \cite{bhatnagar2020ipnet} are also used to penalize the unnatural output. Since AMASS only provides SMPL parameters, we disable the Shape-Comp network and use the results from Motion-Comp network for visualization. Fig. \ref{fig:amass} shows that the proposed method successfully reconstructs the full sequence from such sparse input, which demonstrates the generalization capability of our model to represent novel motions from another data source.

\section{Extended Comparisons to Related Works}
\begin{figure*}[tb]
\centering
\includegraphics[width=1\linewidth]{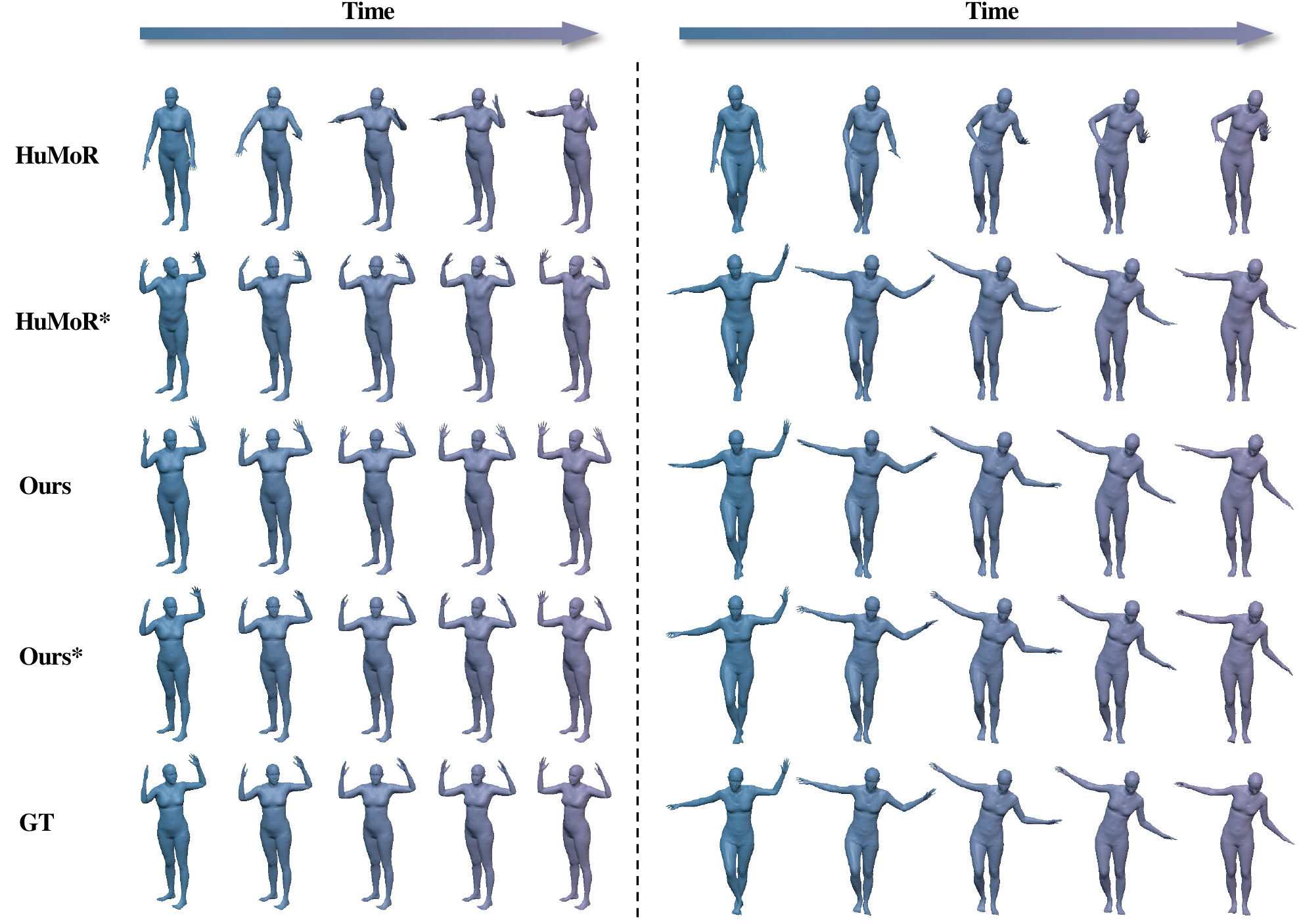}
\caption{\textbf{Qualitative comparisons to HuMoR.} We compare the reconstruction results with HuMoR by using Chamfer loss (HuMoR, Ours) and 3D keypoint loss (HuMoR$^{\ast}$, Ours$^{\ast}$) for auto-decoding. Our method produces more accurate SMPL sequences in both cases.}
\label{fig:humor}
\end{figure*}

\noindent \textbf{Comparison to HuMoR} We compare with a SoTA human body estimation method HuMoR \cite{rempe2021humor} on the task of fitting point cloud sequences. Specifically, we choose 100 mesh sequences of 30 frames from our test set and randomly sample 8192 points from each frame. 
Then we use the pretrained model of both methods to conduct backward optimization with 2 choices of loss functions, \ie Chamfer loss (HuMoR, Ours) or 3D keypoint loss (HuMoR$^{\ast}$, Ours$^{\ast}$). For both losses, we also enabled prior losses to regularize the predicted shape and motion for H4D.
As shown in Tab. \ref{tab:humor}, our method beats HuMoR in both cases. The qualitative comparisons are shown in Fig. \ref{fig:humor}, HuMoR can only recover the global movement trend without accurate limbs by using Chamfer loss and gains a significant improvement when using 3D keypoints as supervision. In contrast, our method obtains more accurate results in both cases.

\begin{table}[htb]
\centering
\small
\renewcommand\tabcolsep{3pt}
	\begin{tabular}{lcccc}
		\toprule[1px]
		& PA-MPJPE $\downarrow$ & MPJPE $\downarrow$ & PVE $\downarrow$ & Accel $\downarrow$ \\
		\midrule
		HuMoR & 70.7 & 46.0 & 45.4 & 10.5 \\
		Ours & 32.6 & 30.0 & 27.6 & 4.9 \\
		\midrule
		HuMoR$^{\ast}$ & 25.7 & 27.1 & 26.1 & 7.3 \\
		Ours$^{\ast}$ & \textbf{16.2} & \textbf{14.5} & \textbf{11.4} & \textbf{4.5} \\
		\toprule[1px]					
	\end{tabular}
	\caption{\label{tab:humor}\textbf{Quantitative comparisons to HuMoR.} By default, the Chamfer loss between the input point cloud and points sampled from the reconstructed mesh is adopted. And $^{\ast}$ indicates that we use 3D keypoint as supervision.}
\end{table}

\begin{figure*}[tb]
\centering
\includegraphics[width=1\linewidth]{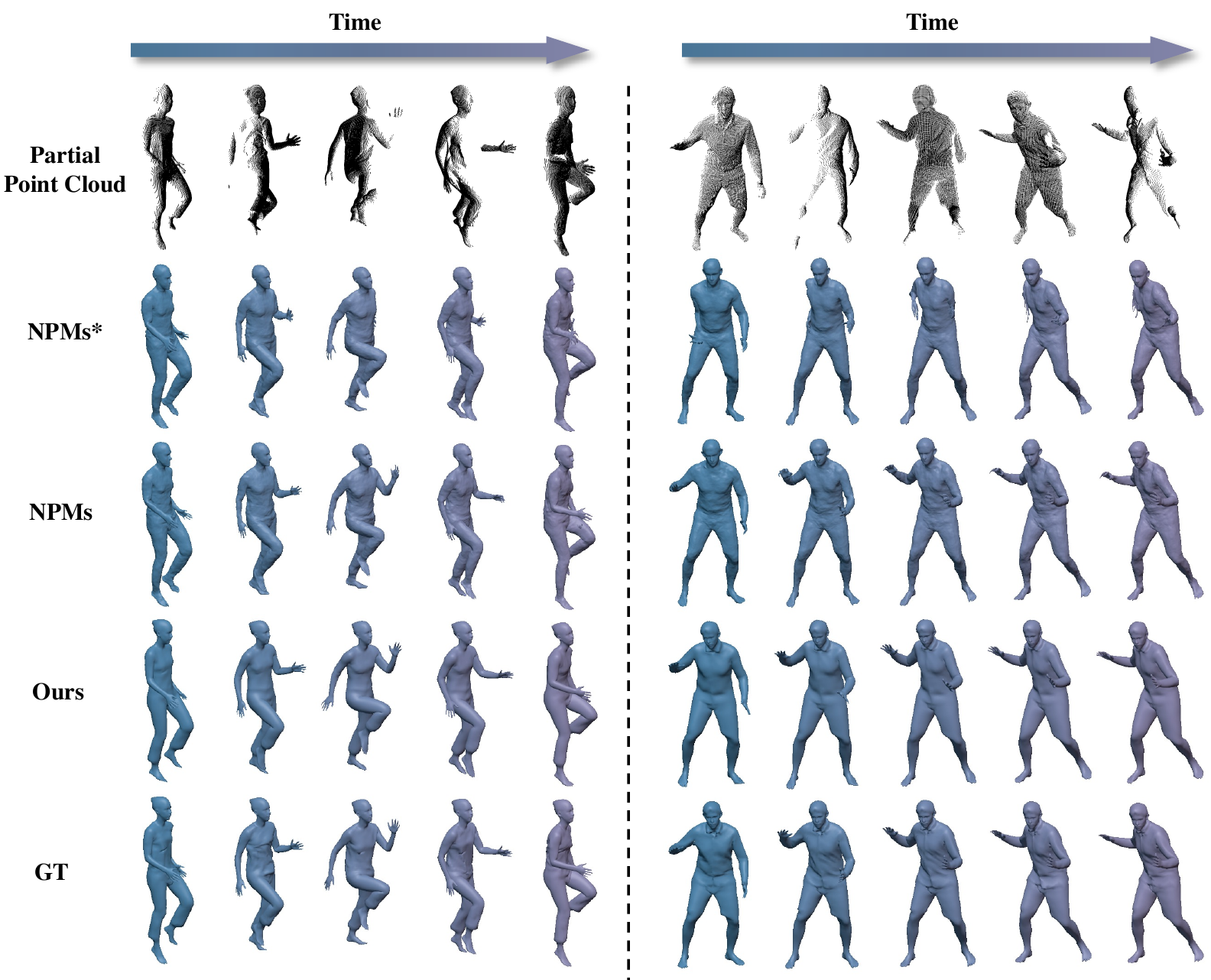}
\caption{\textbf{Qualitative comparisons to NPMs on depth completion task.} We use the partial point clouds back-projected from the depth images rendered by rotating a camera around the performer as observation, and reconstruct the motion sequence with complete geometry. We use the random vectors sampled from a Gaussian distribution $N\left(0,0.01\right)$ to initialize the codes for NPMs$^{\ast}$ and Ours, and use the pretrained encoders to obtain initialization for NPMs.}
\label{fig:npms}
\end{figure*}

\noindent \textbf{Comparison to NPMs} 
We provide the comparisons to NPMs \cite{palafox2021npms} on depth completion task. We choose 100 sequences and use the pretrained model of NPMs to perform completion from partial depth. Specifically, given a depth image sequence of 30 frames, we project the depth values into a $256^3$-SDF grid to generate the inputs for NPMs, and then optimize the latent codes frame-by-frame with the default setup.
Note that NPMs runs 10 times slower than H4D and uses twice of the GPU memory.
Tab. \ref{tab:npms} shows that our model outperforms NPMs, either w/ or w/o (NPMs$^{\ast}$) encoders for code initialization, on both metrics. As can be seen from the qualitative results shown in Fig. \ref{fig:npms}, NPMs produces accurate motion but fails to recover fine-grained geometry, while our results are plausible on both shape and motion.

\begin{table}[htb]
	\centering
	\small
		\begin{tabular}{lcc}
			\toprule[1px]
			&  IoU $\uparrow$ & CD $\downarrow$ \\
			\midrule
			NPMs$^\ast$ & 79.3\% & 0.104 \\
			NPMs & 85.5\% & 0.042 \\
			Ours & \textbf{87.7\%} & \textbf{0.037} \\
			\toprule[1px]
		\end{tabular}
	\caption{\label{tab:npms}\textbf{Quantitative comparisons to NPMs.} We use the random vectors sampled from a Gaussian distribution $N\left(0,0.01\right)$ to initialize the codes for NPMs$^{\ast}$ and Ours, and use the pretrained encoders to obtain initialization for NPMs.}
% 	\vspace{-2in}
\end{table}

\section{Visualization of Principal Components} \label{sec:vis_pca}
\begin{figure*}[htb]
\centering
\includegraphics[width=1\linewidth]{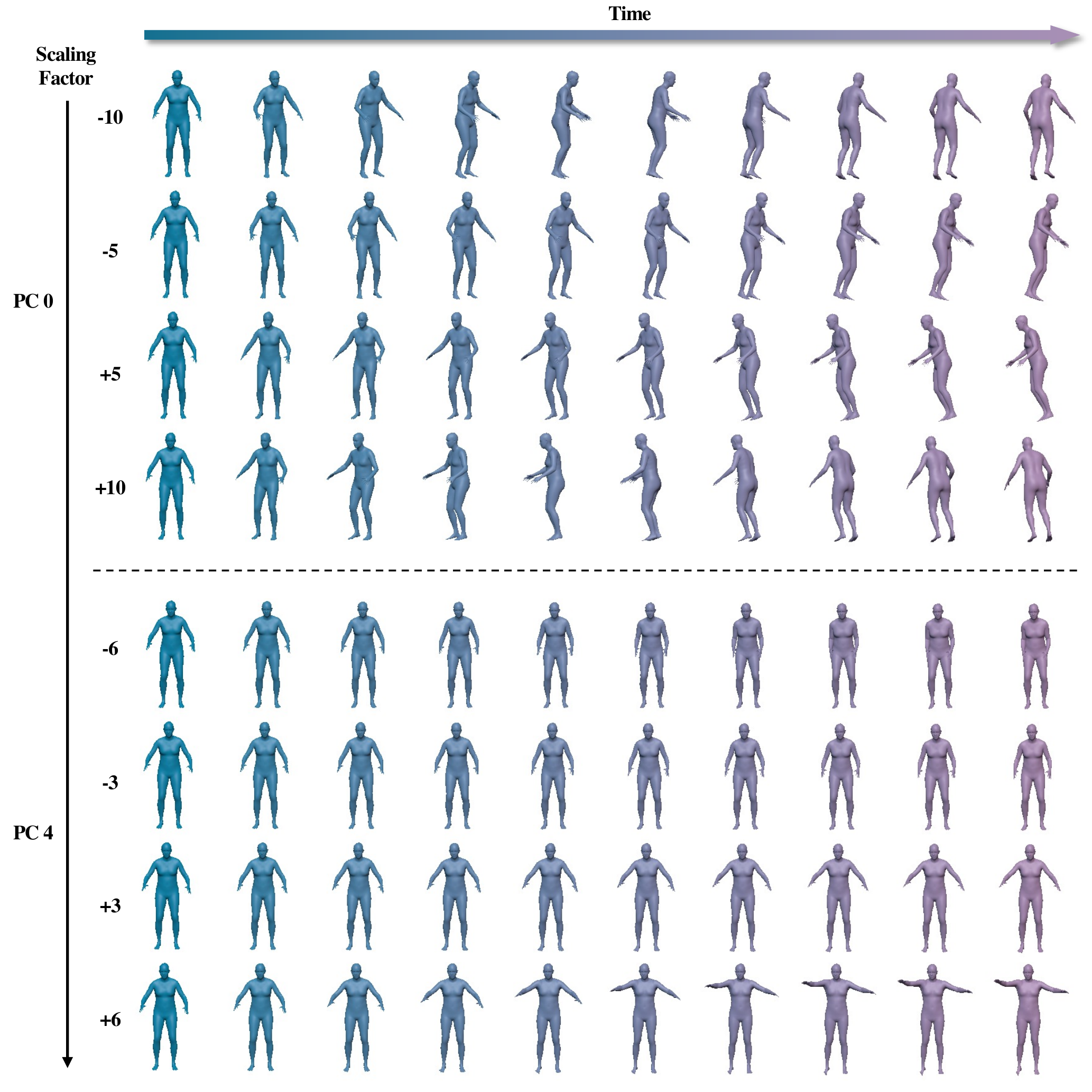}
\caption{\textbf{Visualization of principal components (1).} PC0 and PC4 is the first principal component of global rotation and other body joint rotations respectively. The number before each row is the scaling factor for the corresponding component (multiply it with the eigenvector and show the result motion).}
\label{fig:pca1}
\end{figure*}

\begin{figure*}[htb]
\centering
\includegraphics[width=1\linewidth]{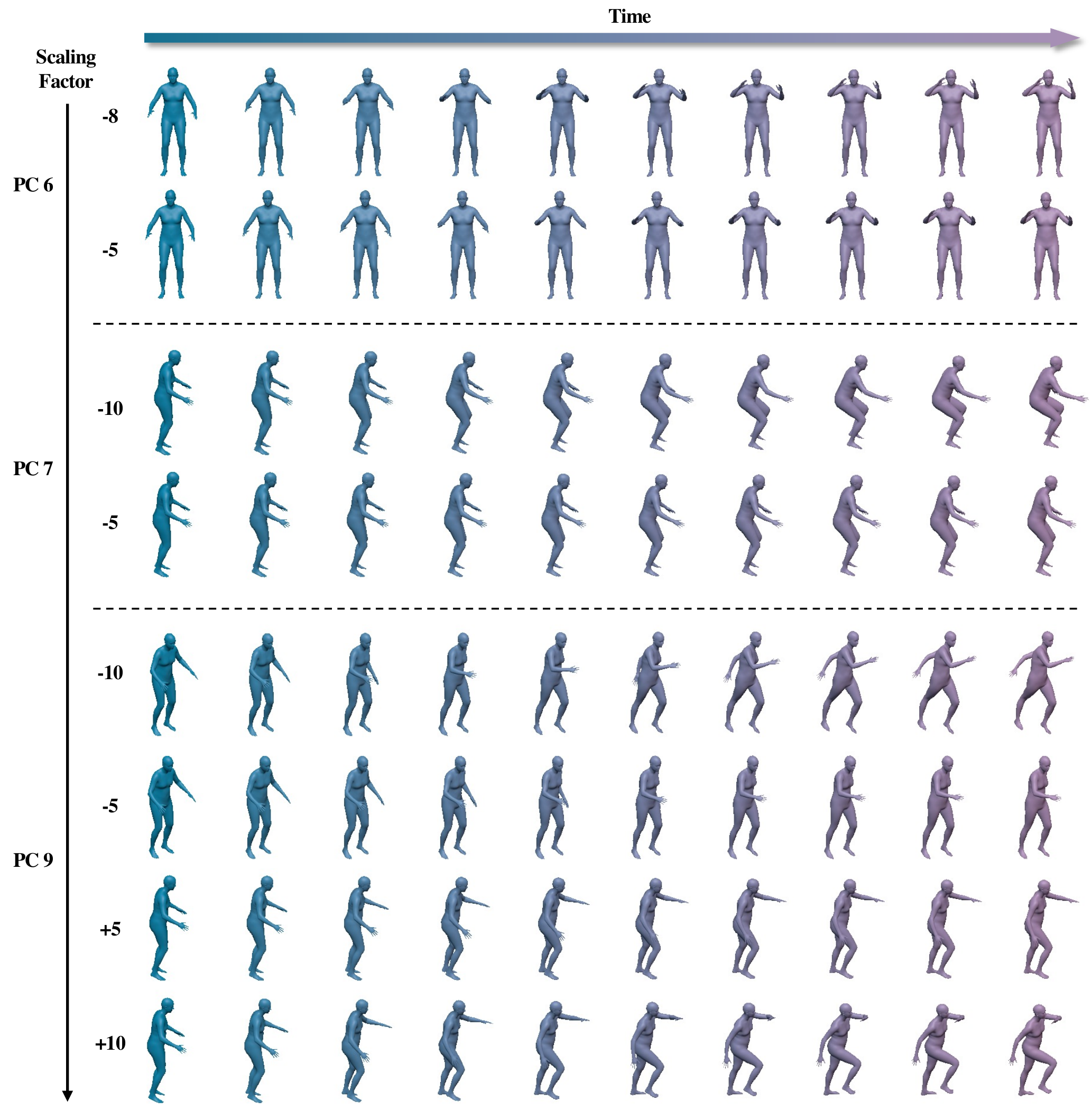}
\caption{\textbf{Visualization of principal components (2).} Here are three principal components of body joint rotations, which in general control the movements of forearms (PC6), the bending of the legs (PC7), and motion similar to running (PC9) respectively.}
\label{fig:pca2} 
\end{figure*}

The linear motion model in our framework totally has 90 principal components, the first 4 components are for global rotation (pelvis joint) and the rest 86 for other body joints. We start from the same rest pose and visualize some principal components in Fig. \ref{fig:pca1} and \ref{fig:pca2}. Specifically, for each shown component, we select different scaling factors (before each row) and multiply them with this component to show the motion results.
As shown, PC0 roughly controls the global rotation around the vertical axis; PC4 and PC6 affect the opening and closing of the upper arm and forearm, respectively; PC7 is related to the bending of the legs; and PC9 tells the movement of arms and legs at the same time. In general, the positive and negative scaling factors of components correspond to opposite directions of motion, and the absolute value affects the magnitude of the motion.

\section{Additional Qualitative Results}
\begin{figure*}[tb]
\centering
\includegraphics[width=0.95\linewidth]{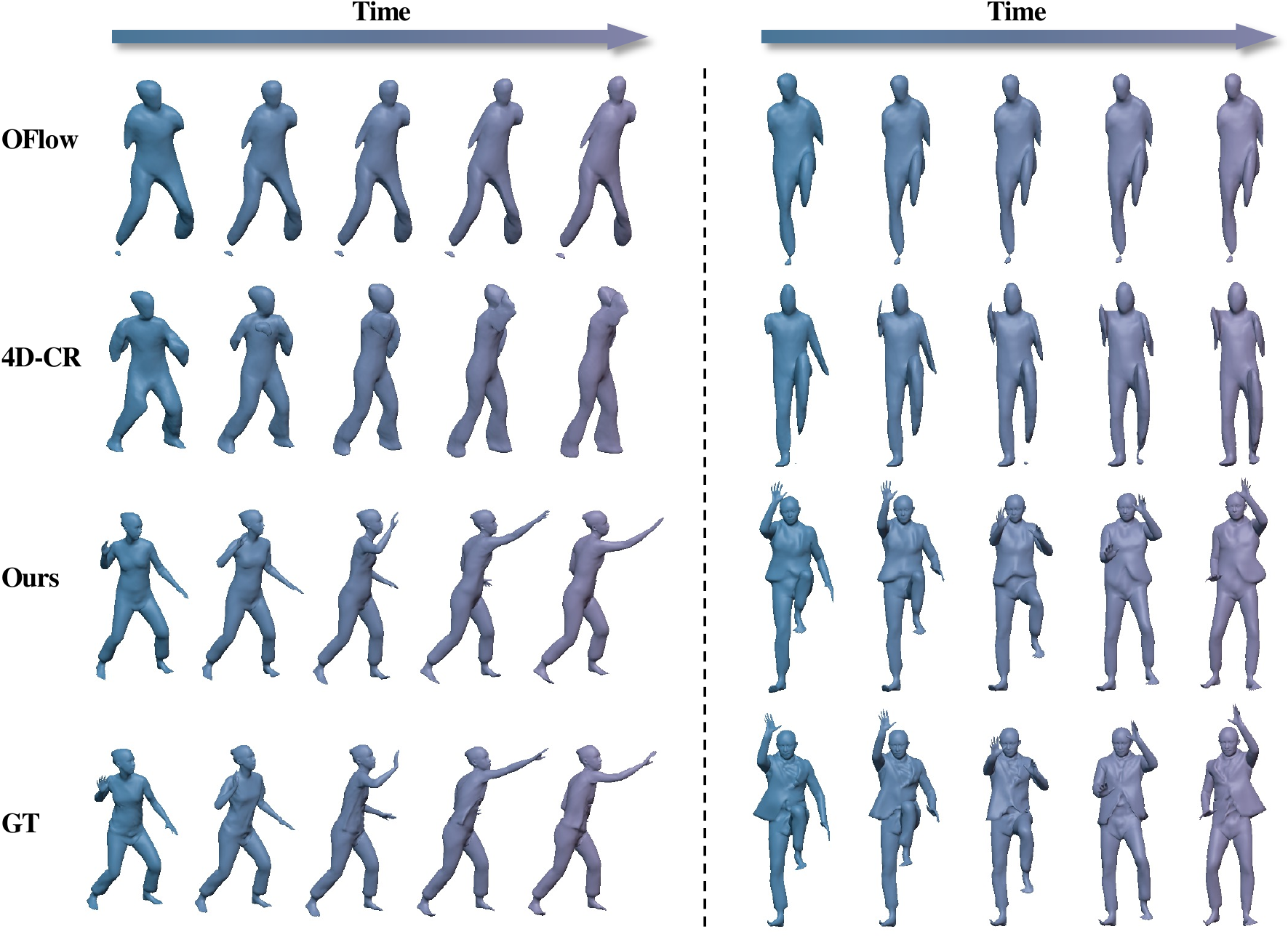}
\caption{\textbf{4D Reconstruction.} Our method produces accurate motion sequence with fine-grained geometry (blazer, long trousers and hairstyle), while the results of baseline methods suffer from incomplete geometry with missing arms or hands, and are overly smooth.}
\label{fig:supp_4d_recons} 
\end{figure*}

\begin{figure*}[tb]
\centering
\includegraphics[width=0.95\linewidth]{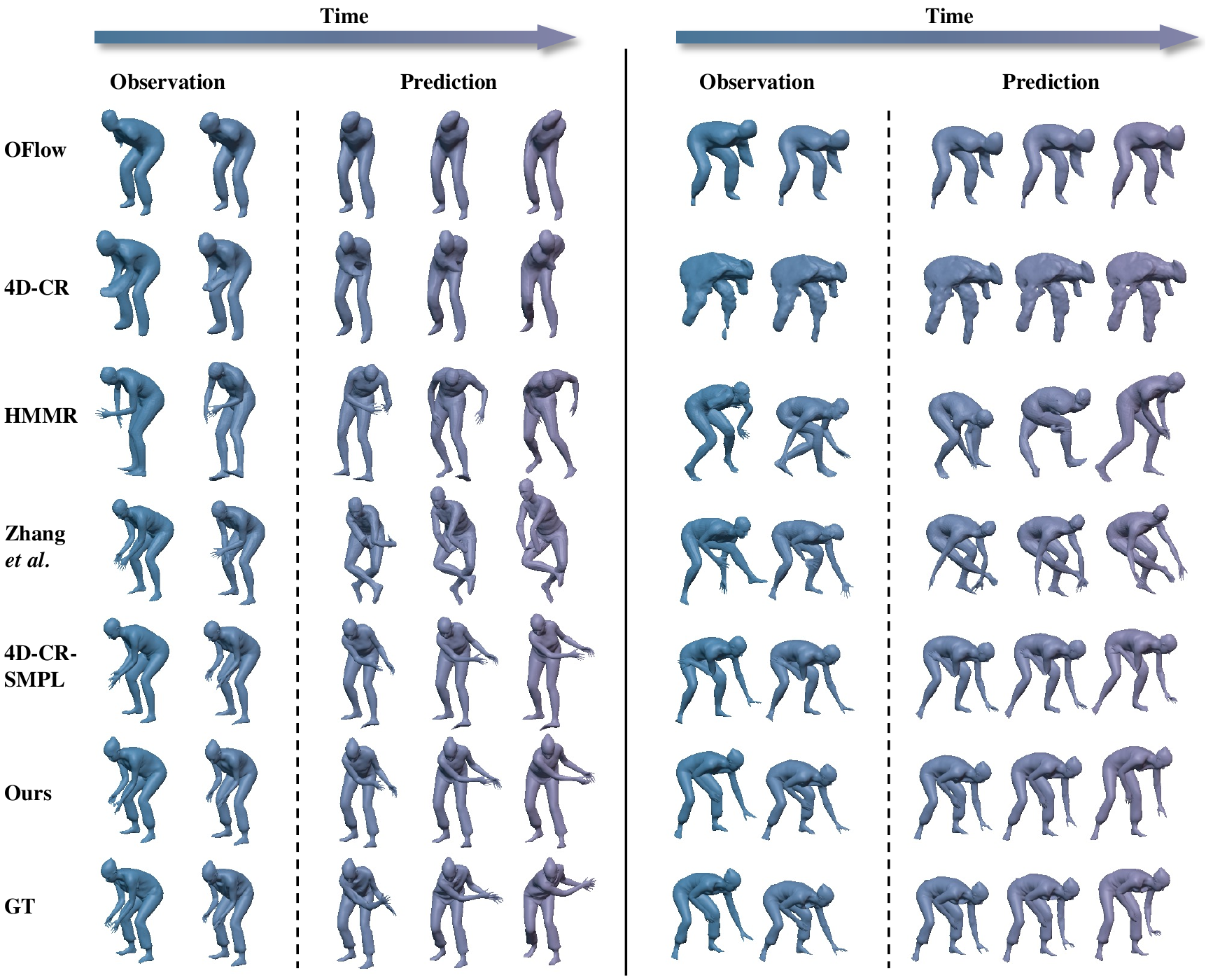}
\caption{\textbf{Future Prediction.} Here are two different sequences on the future prediction task (split by solid line). Each sequence has $L=30$ frames and we are aiming to extrapolate 10 future temporal frames based on 20 past observed frames. The meshes on the left of dotted line are reconstruction results of the observations, and the meshes on the right are the predictions for future time steps.}
\label{fig:supp_future_pred} 
\end{figure*}

\begin{figure*}[tb]
\centering
\includegraphics[width=0.95\linewidth]{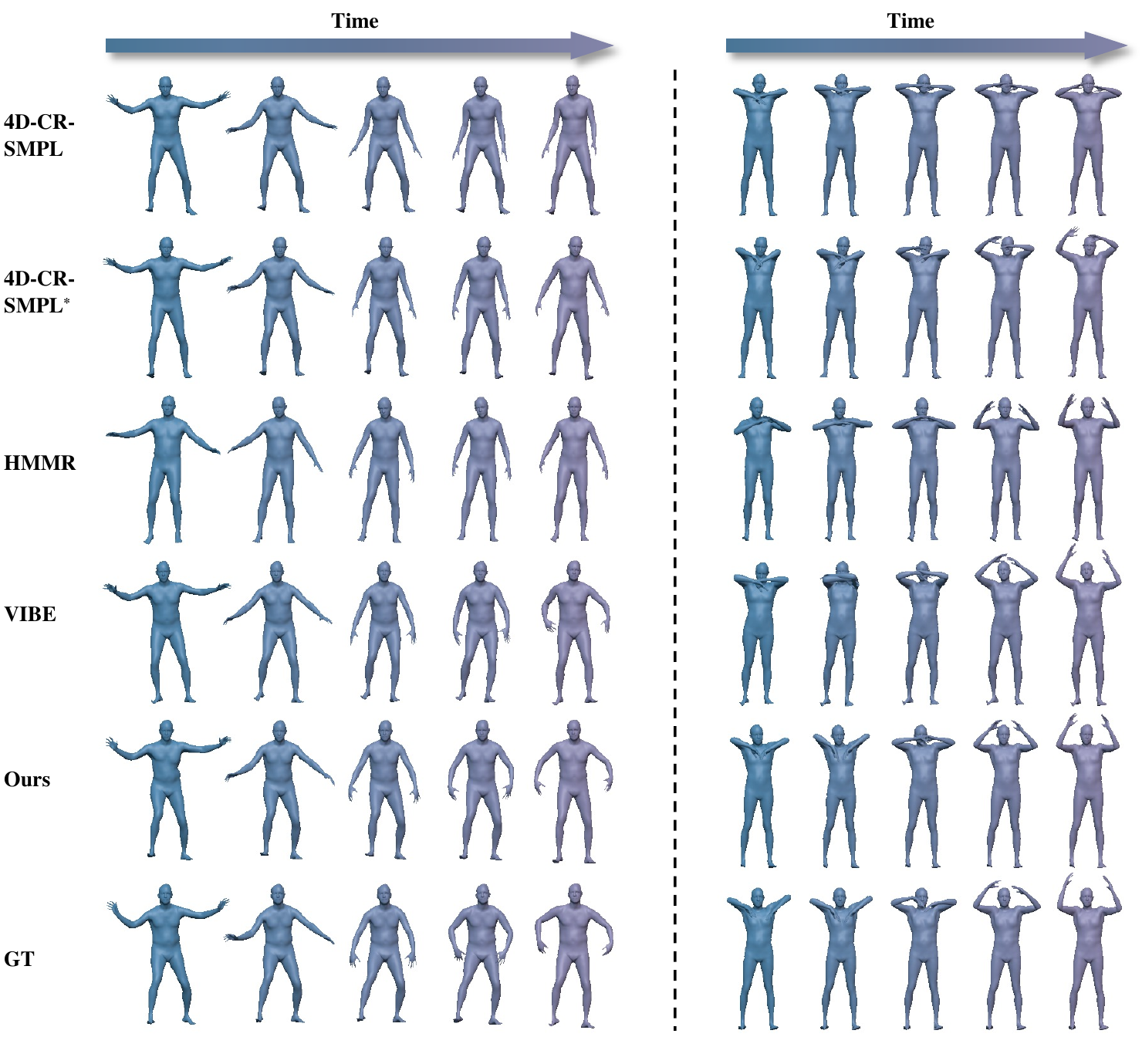}
\caption{\textbf{Shape and Motion Recovery.} Different from the 4D reconstruction task, the goal here is to recover accurate SMPL motion sequence from the input point cloud sequence. We uniformly sample 5 frames (out of 30 frames) for visualization. Our model in general performs best among all the methods.}
\label{fig:motion_recovery} 
\end{figure*}

\begin{figure*}[htb]
\centering
\includegraphics[width=0.9\linewidth]{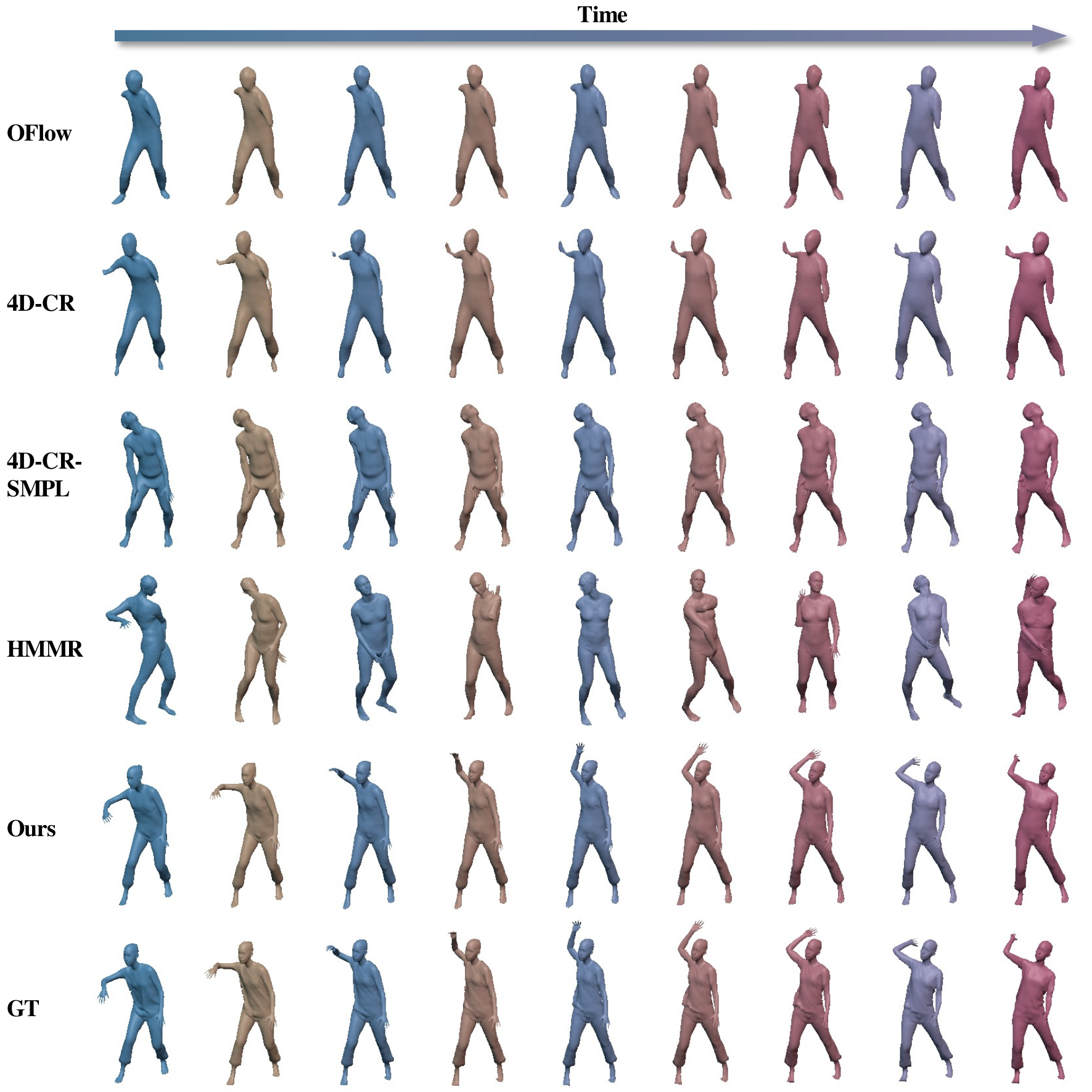}
\caption{\textbf{Temporal Completion.} Given a sequence of $L=30$ frames, we randomly select 15 frames as observation and perform the backward fitting algorithm to optimize the SMPL parameters and latent codes, and then reconstruct the full sequence to complete the missing frames. The meshes with yellow-red-ish color are completed unseen frames. We find 4D-CR-SMPL and HMMR produce unnatural pose results while our method successfully reconstructs and completes the full motion sequence, possibly because our linear motion model provides regularization and global temporal context for the output motion.}
\label{fig:temporal_comp}
\end{figure*}

\begin{figure*}[htb]
\centering
\includegraphics[width=0.9\linewidth]{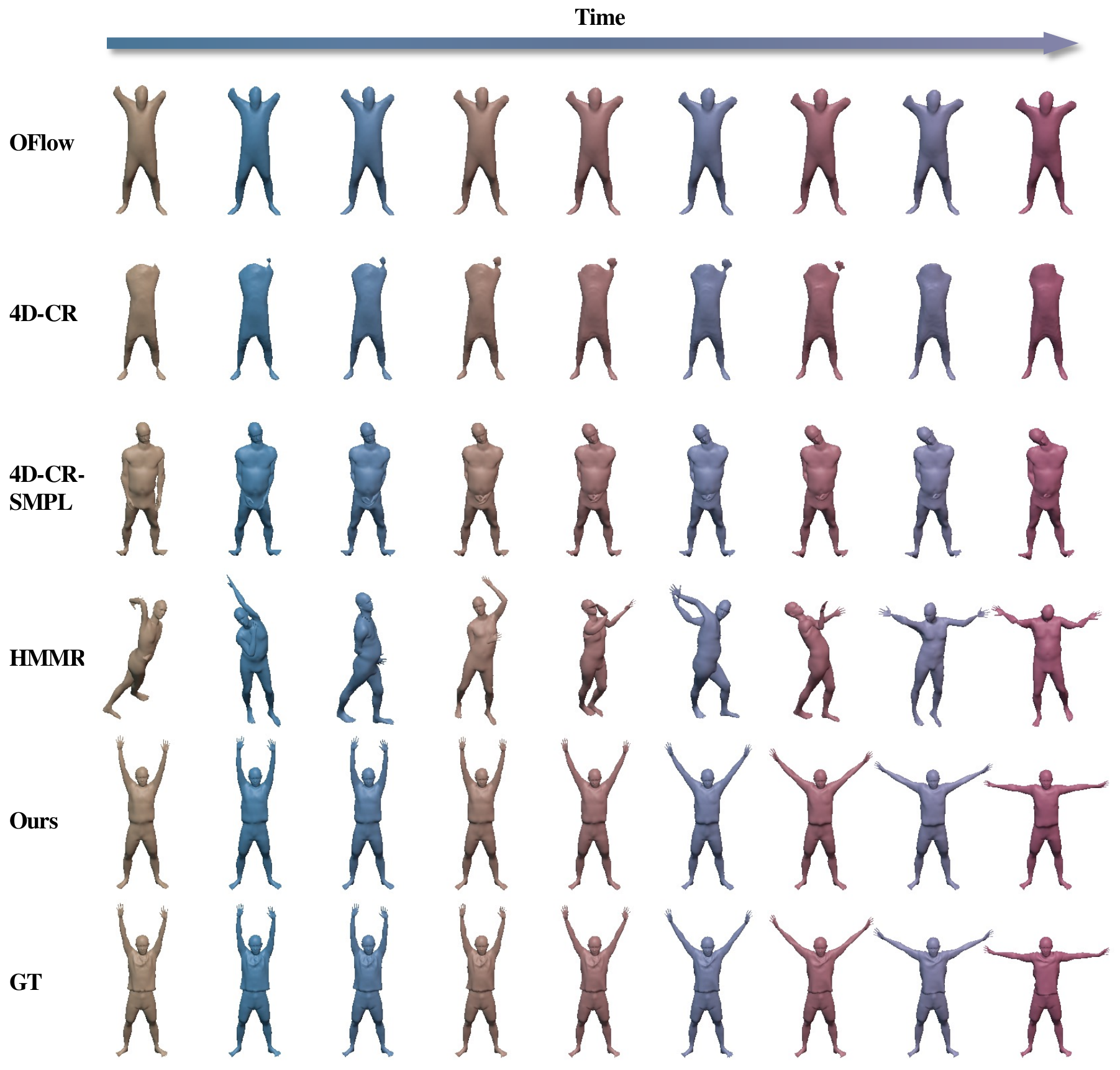}
\caption{\textbf{Temporal Completion.} The meshes with yellow-red-ish color are completed unseen frames.}
\label{fig:temporal_comp2}
\end{figure*}

\begin{figure*}[tb]
\centering
\includegraphics[width=0.9\linewidth]{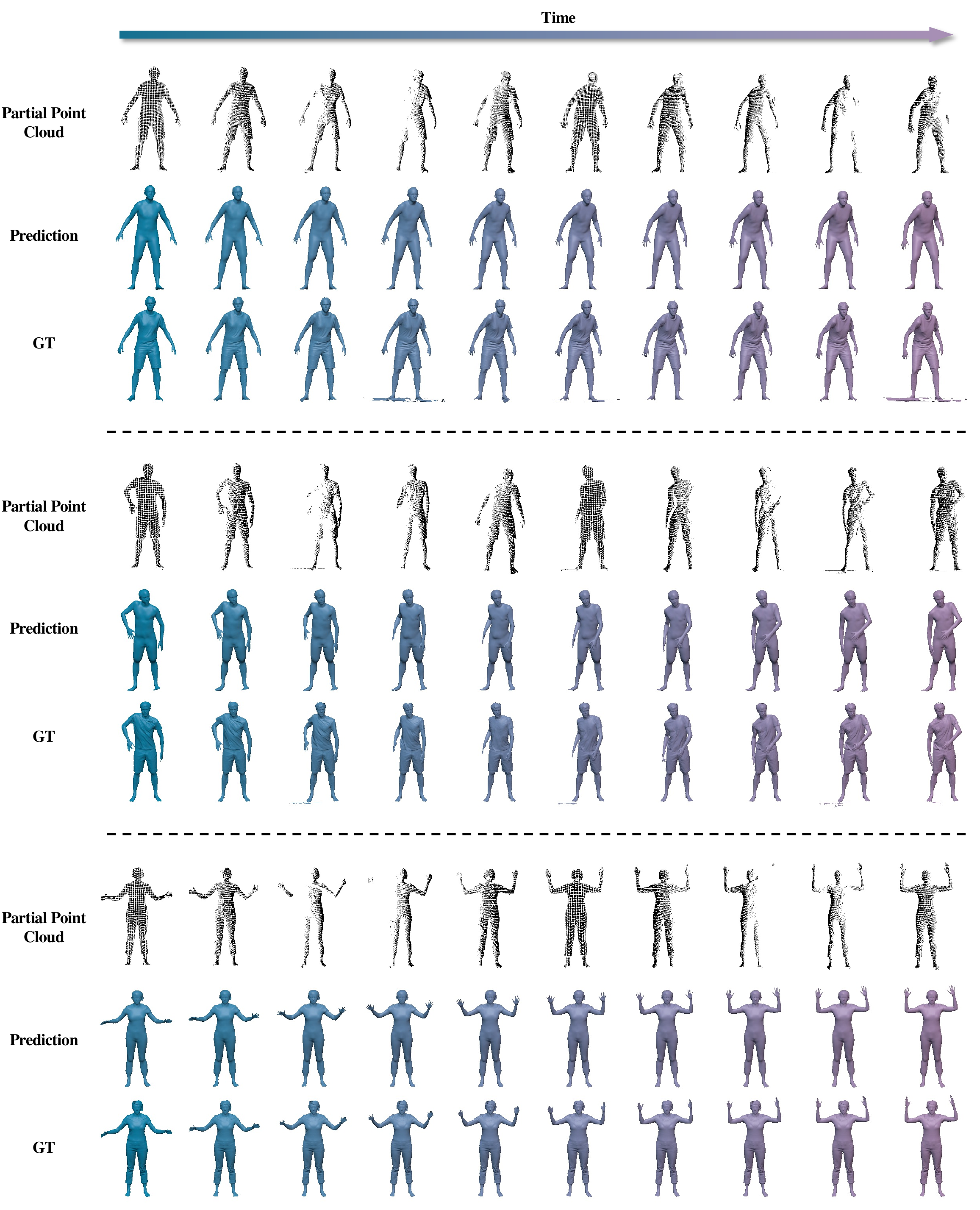}
\caption{\textbf{Spatial Completion.} We use the partial point clouds back-projected from the depth images as observation, and reconstruct the motion sequence with complete geometry. Note that the depth images are rendered from the raw scanned mesh sequence of CAPE dataset, which simulates the real-world scenarios.}
\label{fig:spatial_comp} 
\end{figure*}

\begin{figure}[tb]
\centering
\includegraphics[width=1\linewidth]{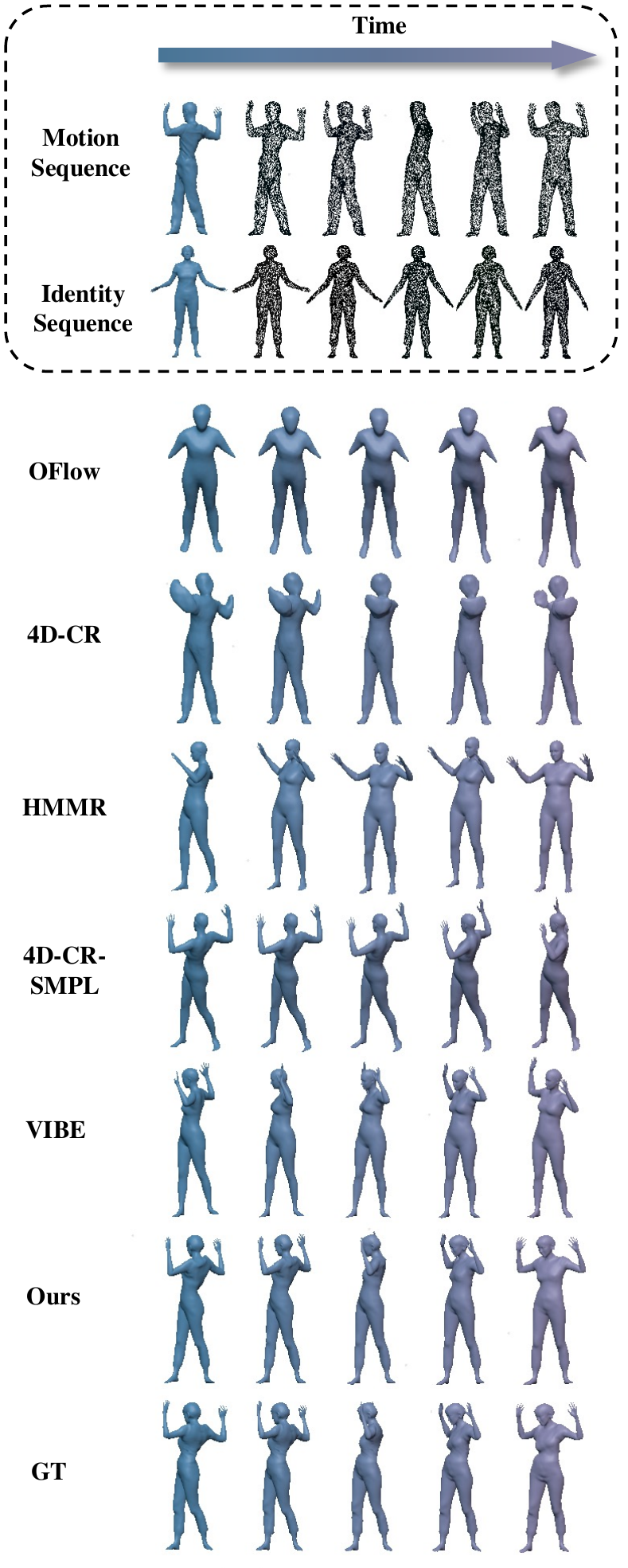}
\caption{\textbf{Motion Retargeting.} Our goal is to transfer the human movements of the motion sequence (Row 1) to the people in the identity sequence (Row 2).}
\label{fig:supp_motion_transfer} 
\end{figure}

\begin{figure}[tb]
\centering
\includegraphics[width=1\linewidth]{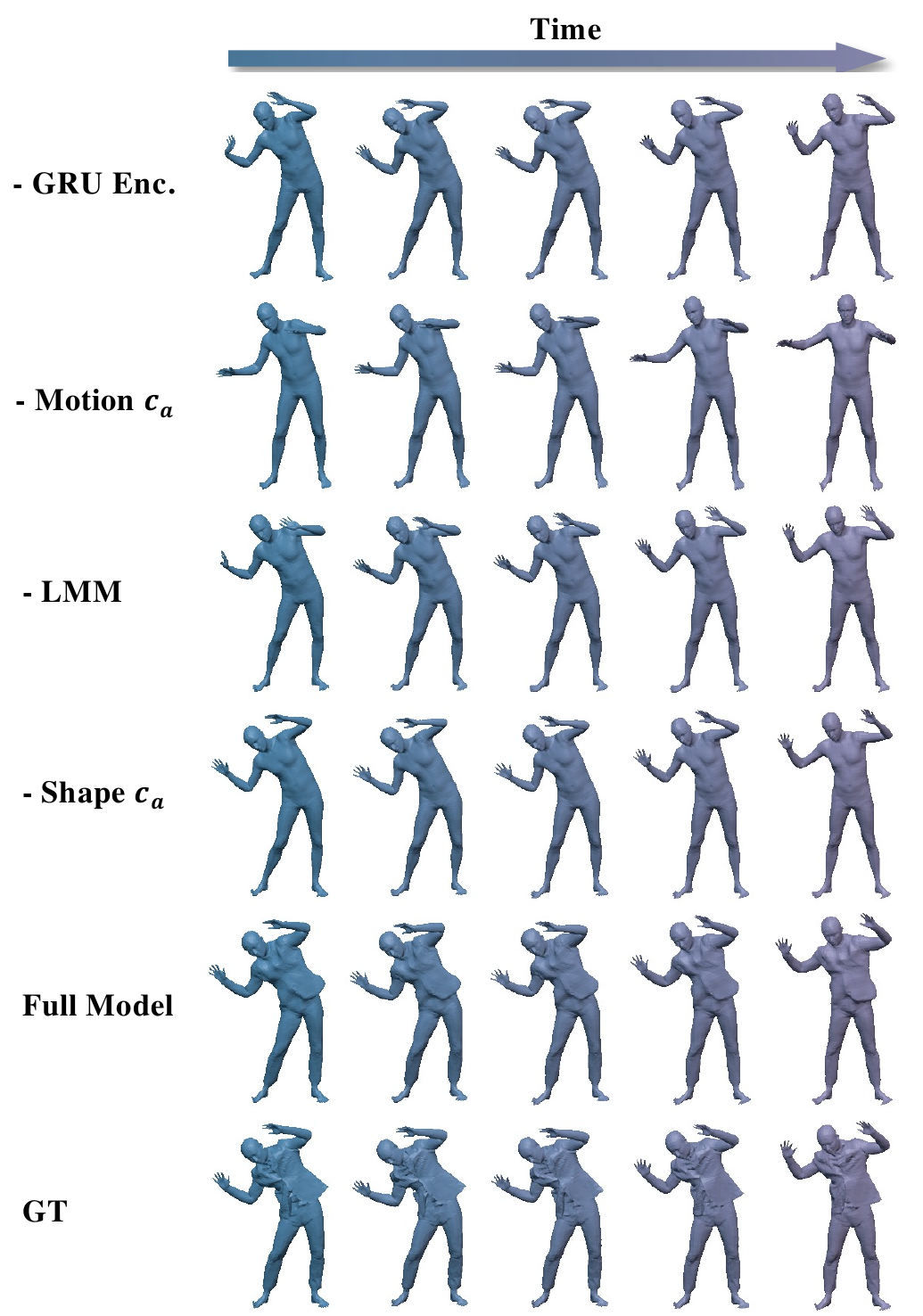}
\caption{\textbf{Ablation Study.}}
\label{fig:ablation_vis} 
\end{figure}

We show additional qualitative examples on 4D reconstruction in Fig. \ref{fig:supp_4d_recons}, shape and motion recovery in Fig. \ref{fig:motion_recovery}, temporal completion in Fig. \ref{fig:temporal_comp} and \ref{fig:temporal_comp2}, spatial completion in Fig. \ref{fig:spatial_comp}, future prediction in Fig. \ref{fig:supp_future_pred}, motion retargeting in Fig. \ref{fig:supp_motion_transfer} and ablation study in Fig. \ref{fig:ablation_vis}.

\section{Run-time}
In Tab. \ref{tab:runtime}, we show the per sequence run-time of our method and previous 4D representation methods on forward inference for 4D reconstruction and backward optimization for temporal completion. Note that we report the time cost to run a complete backward optimization process for a sequence (500 iterations). The length of the full sequence is $L=30$, and all models run on a single NVIDIA 2080Ti GPU. Instead of the Neural ODE \cite{chen2018neural}, we model the human motion using the linear model and GRU-based compensation networks. As can be seen, our model runs faster in both cases, especially in backward optimization.

\begin{table}[htb]
    \centering
    \small
    \begin{tabular}{lcc}
     \toprule[1px]
     & Forward (s) & Backward (min)\\
     \midrule
    OFlow \cite{niemeyer2019occupancy} & 1.106 (0.814) & 17.600\\
    4D-CR \cite{jiang2021learning} & 14.469 (5.861) & 14.117\\
    4D-CR-SMPL \cite{jiang2021learning} & 0.209 & 16.817\\
    Ours & \textbf{0.175} & \textbf{7.303}\\
     \toprule[1px]
    \end{tabular}
    \vspace{-0.1in}
\caption{\label{tab:runtime}\textbf{Comparisons about the run-time.} We show per sequence run-time of our method and baselines on forward inference and backward optimization. The numbers in the parentheses mean time without Marching Cubes.}
\vspace{-0.1in}
\end{table}

\section{Limitations and Future work}
We now discuss a few limitations of our approach that point to future work.
\textbf{First}, our motion model can reconstruct the discretized frame w.r.t each input time step, but not the arbitrary time in the continuous whole time span like 4D-CR \cite{jiang2021learning} or OFlow \cite{niemeyer2019occupancy}, which will be useful in some scenarios requesting higher temporal resolution from inputs. Incorporating a network that takes a time value scalar as input, \eg Neural ODE \cite{chen2018neural}, temporal MLP, would be a solution.
\textbf{Second}, we currently conduct all experiments on the sequences of 3D data, \eg point clouds or meshes. 
On the one hand, this is due to the lack of 4D human datasets with color images, \eg pairs of video and 3D human sequences (with clothing and hair). And on the other hand, the focus of this work is to propose a compositional representation and effectively power various 4D human-related applications based on point cloud. Combining our representation with techniques such as neural rendering \cite{liu2019softras,mildenhall2020nerf} or photometric-based optimization \cite{lin2019photometric,liu2020dist} for image-based full human 4D reconstruction would be a promising future direction.
\textbf{Third}, we adopt the same clothing representation used in previous work \cite{alldieck2019tex2shape,lazova2019360,CAPE:CVPR:20}, \ie per-vertex offsets upon the body in the canonical pose, and extend it to apply to temporal sequences. However, as discussed in CAPE \cite{CAPE:CVPR:20}, some loose garments such as skirts and coats are difficult to represent with offsets due to the limited capacity. Modeling clothes and hair as separate layers on the body with meshes or implicit surface is a feasible way, and we leave it to future work.
\textbf{Forth}, since we have a compact motion representation that uses one single motion code to provide global control upon the whole sequence, future works also include high-level inference applications such as using the motion code learned in an unsupervised fashion to perform action classification with a simple linear classifier.

\section{Broader Impact and Social Impact}
Learning a compact representation for 3D data is a widely interested problem. However, less attention has focused on the 4D cases, though it is important for various applications to understand time-varying objects, \eg Robotics, VR/AR. This work focuses on 4D human modeling and proposes H4D, a compact and compositional representation, which uses low-dimensional SMPL parameters and latent codes to encode key factors of dynamic humans. We make some attempts and demonstrate our representation has rich capacity and is amenable to many applications. We hope these explorations could provide insights for future research directions. For instance, using our representation for video-based full human reconstruction; exploiting the compositional property to control the outputs for generative tasks; and improving the 4D human representation and make up for the discussed limitations of our method. Broadly, our approach can serve as an important core tech in achieving the Metaverse. It may enable everyone to produce their own Avatar with their motions, potentially benefiting the Social Welfare.

\end{document}